\documentclass[a4paper,12pt]{article}

\usepackage{authblk} % Für die Affilitions
\usepackage{graphicx} % Um Plots einzufügen
\usepackage{booktabs} % \toprule und so
\usepackage{xcolor} % Farbige Schrift
\usepackage{tabularx}
\usepackage{array}
\usepackage{threeparttable}
\usepackage{threeparttablex}
\usepackage{multirow} % \multicolumn
\usepackage[normalem]{ulem} % Durchgestrichene Wörter
\usepackage{biblatex} %Imports biblatex package
\usepackage{caption}
\usepackage[breaklinks]{hyperref}
\usepackage{xurl}
\providecommand{\keywords}[1]
{
  \small
  \textbf{\textit{Keywords---}}{#1}
}

\begin{document}

\title{Backdoor attacks on DNN and GBDT - A Case Study from the insurance domain}

\author[1]{Robin K\"uhlem}

\author[1]{Daniel Otten}

\author[1]{Daniel Ludwig}

\author[1,3]{Anselm Hudde}

\author[2]{Alexander Rosenbaum}

\author[2]{Andreas Mauthe}

\affil[1]{Debeka, Koblenz, Germany}

\affil[2]{Computer Science, University of Koblenz, Koblenz, Germany}

\affil[3]{Department of Maths and Technology, Koblenz University of Applied Sciences, Remagen, Germany}

%\presentaddress{This is sample for present address text this is sample for present address text.}

%\fundingInfo{Text}
%\JELinfo{ejlje}

\maketitle

\begin{abstract}
Machine learning (ML) will likely play a large role in many processes in the future, also for insurance companies. However, ML models are at risk of being attacked and manipulated. In this work, the robustness of Gradient Boosted Decision Tree (GBDT) models and Deep Neural Networks (DNN) within an insurance context will be evaluated.  Therefore, two GBDT models and two DNNs are trained on two different tabular datasets from an insurance context. Past research in this domain mainly used homogenous data and there are comparably few insights regarding heterogenous tabular data. The ML tasks performed on the datasets are claim prediction (regression) and fraud detection (binary classification). For the backdoor attacks different samples containing a specific pattern were crafted and added to the training data. It is shown, that this type of attack can be highly successful, even with a few added samples. The backdoor attacks worked well on the models trained on one dataset but poorly on the models trained on the other. In real-world scenarios the attacker will have to face several obstacles but as attacks can work with very few added samples this risk should be evaluated.
\end{abstract}


% \renewcommand\thefootnote{}
% \footnotetext{\textbf{Abbreviations:} ANA, anti-nuclear antibodies; APC, antigen-presenting cells; IRF, interferon regulatory factor.}

% \renewcommand\thefootnote{\fnsymbol{footnote}}
% \setcounter{footnote}{1}

\section{Introduction} \label{secIntroduction}

Utilizing machine learning (ML) has great potential in many domains, but especially for insurance companies \cite{Eling2022}. McKinsey \cite{Balasubramanian2021} and Deloitte \cite{Deloitte2017} predict a fundamental impact of Artificial Intelligence (AI) on the insurance industry. Large shifts are expected till the end of this decade, and insurers must evaluate the use of AI \cite{Balasubramanian2021} since it can provide significant benefits in several areas for insurance companies \cite{Deloitte2017,Balasubramanian2021}. Consequently, not taking advantage of these benefits puts companies in danger of falling behind the competition.
Automating and making processes more efficient will help to save costs and provide better customer service through decreased response times \cite{Balasubramanian2021}. Concrete use cases include underwriting, fraud detection, customer service, and many other tasks \cite{Deloitte2017}. For example, in the case of underwriting, automation via AI can save costs and lead to more accurate decisions while enabling faster offers to customers and more individualized premiums \cite{Eling2022}.
The downside is that the usage of ML does not come without risks \cite{Xue2020}. ML models have several vulnerabilities that attackers can exploit \cite{Xue2020}. Given these vulnerabilities, companies must assess the risks and mitigate weaknesses before heavily relying on ML in their business processes. Yet, while companies are aware of the importance of ML security, many do not take appropriate measures \cite{Kumar2020}. For example, if the insurance underwriting process is fully automated and performed by ML models, a successful attack could mean that insurance premiums are set much lower than appropriate, and the insurer loses money on those contracts. In another instance, attacking an ML-based fraud detection can allow attackers to get compensation for fraudulent claims. Consequently, reaping the mentioned benefits of the application of ML comes with certain risks for insurance companies.

In this work the effect of backdoor attacks on Deep Neural Networks (DNNs) and Gradient Boosting Decision Tree (GBDT) models is examined. Both ML algorithms provide good performance on tabular data \cite{Borisov2022,Shwartz-Ziv2021} and are therefore both viable options for many insurance use cases. As a foundation for the experiments two tabular datasets from an insurance context were used. AI systems should have a certain degree of trustworthiness to enable their full potential in real-world applications \cite{Kaur2022}, which makes research regarding their vulnerabilities highly relevant. This is especially true regarding (heterogenous) tabular data, as most past research in this domain focused on homogenous data \cite{Sadeghi2020}.

The remainder of this paper is structured as follows. Firstly, the theoretical background will be explained and next the related work will be described. Subsequently, the data used will be elucidated in Chapter \ref{secDatasets} and data preprocessing and the training process of the models in Chapter \ref{secModelsPreprocessing}. The backdoor attacks and the corresponding results are described in Chapter \ref{secBackdoorAttacks}. Chapter \ref{secDiscussion} contains a discussion of the results and there is a conclusion regarding the most important findings of this work in Chapter \ref{secConclusion}. Lastly, implications, directions for future research and limitations are summarised in Chapter \ref{secFutureWorkLimitations}.

\section{Theoretical Background on ML Attacks} \label{secTheoreticalBackground}

The attacks on ML systems can be divided into attacks during the test phase and attacks during the training phase \cite{Xue2020}. Attacks during the training phase include data poisoning and backdoor attacks \cite{Xue2020}. Data poisoning refers to the insertion of malicious data to alter the performance from the model \cite{Xue2020}. Backdoor attacks can be conducted through alteration of model parameters \cite{Ji2017} or data poisoning \cite{Chen2017b}. During a backdoor attack through data poisoning, specific data are added to the training data to implement a backdoor in the model \cite{Chen2017b}. This enables the attacker to achieve the desired output of the model by adding a certain pattern to the inputs \cite{Chen2017b,Joe2022}.
Attacks during the test phase include Adversarial Example (AE) attacks, model stealing, and recovery of training data \cite{Xue2020}. In an AE attack, small perturbations are added to samples to change the classification of these samples \cite{Zhang2020a}. In a model stealing (model extraction) attack, the adversary uses the output of a model to reconstruct it \cite{Xue2020}. In the case of recovery of training data, the attacker can infer whether certain data have been used for training or determine certain aspects of the training data \cite{Xue2020}.

Attacks during the training phase manipulate the training data to achieve the attacker's goals \cite{Liu2018}. Consequently, the attacker must have the ability to alter the training data, which can be difficult since this data should be well protected in a real-world setting \cite{Liu2018}. Easier attack approaches are the implementation of adversarial samples in the training data, for instance when the model is retrained to keep it up to date \cite{Liu2018} or when data from external, potentially uncontrolled sources are used \cite{Koh2022}. Even a small fraction of adversarial samples in the training data can have a large impact \cite{Xue2020}. Therefore, the integrity of the training data must be ensured \cite{Schwarzschild2021} and all parts of the system must be secure to enable a high level of security of the ML application \cite{Schneier2020}.
Data poisoning attacks can be triggerless or backdoor attacks \cite{Schwarzschild2021}. Triggerless attacks make the model classify a specific sample or group of samples falsely without containing a specific trigger \cite{Schwarzschild2021}. For this purpose, samples can be systematically manipulated and added to the training set \cite{Zhu2019}. However, the most popular approach for backdoor attacks works with the addition of samples in the training data that contain a certain pattern \cite{Goldblum2023}. If this pattern / trigger is included in a sample at test time, it will likely lead to the model classifying it as the desired class \cite{Chen2017b}. In view of, for example, a fraud detection model, this would give the attacker the power to have every sample that includes this pattern labeled as non-fraudulent. This opens the door to large-scale insurance fraud.

Backdoor attacks have been proven to work against several different ML algorithms \cite{Huang2022,Weber2023}. Past research on backdoor attacks on ML algorithms focused on settings with homogenous data \cite{Li2022a}. Some previous works that used tabular data made use of a temporal structure in the data \cite{Joe2021}, missing value patterns \cite{Joe2022} or a substitute dataset to implement a backdoor through fine-tuning a model to construct the backdoors \cite{Lv2023}. Another backdoor attack was developed based on a malware classifier \cite{Li2022b}. Certain features of the malware were changed systematically to evade detection. The way in which features could be changed was restricted to keep the malicious samples functioning and realistic \cite{Li2022b}. Restrictions in feature modifications are a common issue with tabular data as categorical values can only take on certain values, bounded value ranges for numerical features, and dependencies between features. Backdoor attacks can evade detection more easily compared to other data poisoning attacks since they ideally do not affect the general performance of the model \cite{Joe2021}. While some proposed attacks perform worse under real-world conditions, they still pose a clear threat \cite{Schwarzschild2021}.

Since previous research in the domain of ML security has mainly focused on homogenous data (consequently also not focuding on tabular data) \cite{Sadeghi2020}, using tabular data from the insurance field with corresponding ML tasks to analyze the security of DNNs and GBDT is a new research approach. It will generate insights regarding the security of ML applications in this domain.

To allow for a comparison between different works in the field of adversarial machine learning, Sadeghi et al. \cite{Sadeghi2020} developed a framework that will be used in this paper. It will enable other researchers to repeat the experiments and provides a clear description of the settings of the experiments \cite{Sadeghi2020}. Firstly, the used ML models are defined, including the ML algorithm, the dataset, and the baseline performance of each model. Secondly, the attack model is analyzed. This includes the knowledge of the adversary, which encompasses aspects regarding the data and the algorithm. Thirdly, the capabilities of the attacker are defined, which mainly refers to the attacker's extent of ability to alter training or testing data. Fourth, the goal of the attacker is defined based on the CIA triad that includes Confidentiality, Integrity, and Availability. Confidentiality is lost when sensitive information is retrieved by an unauthorized third party. For example, a loss of integrity in ML applications can mean that an ML model misclassifies malicious samples. Loss of availability is present when regular samples are misclassified to an extent that hinders normal system operation \cite{Sadeghi2020}. The goals can each additionally be further classified as indiscriminate or targeted \cite{Sadeghi2020}.

Lastly, the strategy of the adversary is identified. This refers to the precise approach the adversary chooses to reach his goals. The attacks can be categorized as active or passive \cite{Sadeghi2020}. Passive attacks only observe the system without altering the normal operations \cite{Nasr2019}. In contrast, active attacks alter the normal behavior of the system \cite{Sadeghi2020}. Furthermore, active attacks can be differentiated into test-time or training-time attacks \cite{Sadeghi2020}.

\section{Datasets} \label{secDatasets}

As a foundation for the training of the models two different datasets from the insurance domain were used. The first dataset is from the health insurance and the second dataset from the vehicle insurance domain. The underlying task for the ML models is a claim prediction (regression) for the health insurance dataset and a fraud detection (binary classification) for the vehicle insurance dataset.

\subsection{Health Insurance Dataset}

Table 1 shows the columns with the corresponding feature types and values of the health insurance dataset (HID). The dataset has been retrieved from Kaggle \cite{Gupta2017}.  The HID has 15,000 entries and 13 columns. There are 396 missing values in the column ``age,'' 956 missing values in the column ``bmi,'' and 756 missing values in the column ``blood pressure'' which is relevant regarding the data preprocessing for the training of the models. Apart from that, there are no missing values. The dataset contains duplicates; after the removal of duplicates, there are 13,904 remaining entries. ``claim'' is the target feature and refers to the amount of insurance claims each insurant made.

\begin{center}
\begin{table*}[!h]%
\caption{Columns, types, and unique values of HID\label{tab1}}
\fontsize{10pt}{14}\selectfont
\begin{tabular*}{\textwidth}{@{\extracolsep\fill}lll@{}}
\toprule
  Columns & Type & Values \\
\midrule
  age & numeric & 18 to 64 (years), NaN \\
  sex & categorical & female, male \\
  weight & numeric & 34 - 95 \\
  bmi & numeric & 16 to 53.1, NaN \\
  hereditary diseases & categorical & NoDisease, Epilepsy, EyeDisease, Alzheimer, Arthritis, \\
  & & HeartDisease, Diabetes, Cancer, High BP, Obesity \\
  no of dependents & numeric & 0 to 5 \\
  smoker & categorical & 0 or 1 \\
  city & categorical & 91 different US cities \\
  blood pressure & numeric & 40 to 122, NaN \\
  diabetes & categorical & 0 or 1 \\
  regular exercise & categorical & 0 or 1 \\
  job title & categorical & Actor, Engineer, Academician, Chef, HomeMakers, \\
  & & Dancer, Singer, DataScientist, Police, Student, \\
  & &  Doctor, Manager, Photographer, Beautician, CA, \\
  & & Blogger, CEO, Labourer, Accountant, FilmDirector, \\
  & & Technician, FashionDesigner, Architect, HouseKeeper, \\
  & & FilmMaker, Buisnessman, Politician,\\ 
  & & DefencePersonnels, Analyst, Clerks, ITProfessional, \\
  & & Farmer, Journalist, Lawyer, GovEmployee \\
  claim & numeric & 1,122 to 63,770 \\
\bottomrule
\end{tabular*}
\end{table*}
\end{center}

Regarding the blood pressure, it appears likely that the values refer to the diastolic blood pressure. This is the most reasonable conclusion in relation to the reference values \cite{Whelton2022}. The original source of the data is unclear. Therefore, no further information could be obtained.

One reason for the choice of this dataset is the health insurance domain. Under normal circumstances a slight change of features like weight or height is possible without raising suspicion. In contrast, in the case of vehicle insurance, changing a car's weight can easily be proven as a false statement. Additionally, many correlations between features and the target variable are intuitive, which makes it easier to develop experiments and comprehend the results.

\begin{figure*}[t]
  \vspace{-20pt}  % reduces blank space above figure, needed when figure is at top of page
  \centerline{\includegraphics[width=30pc]{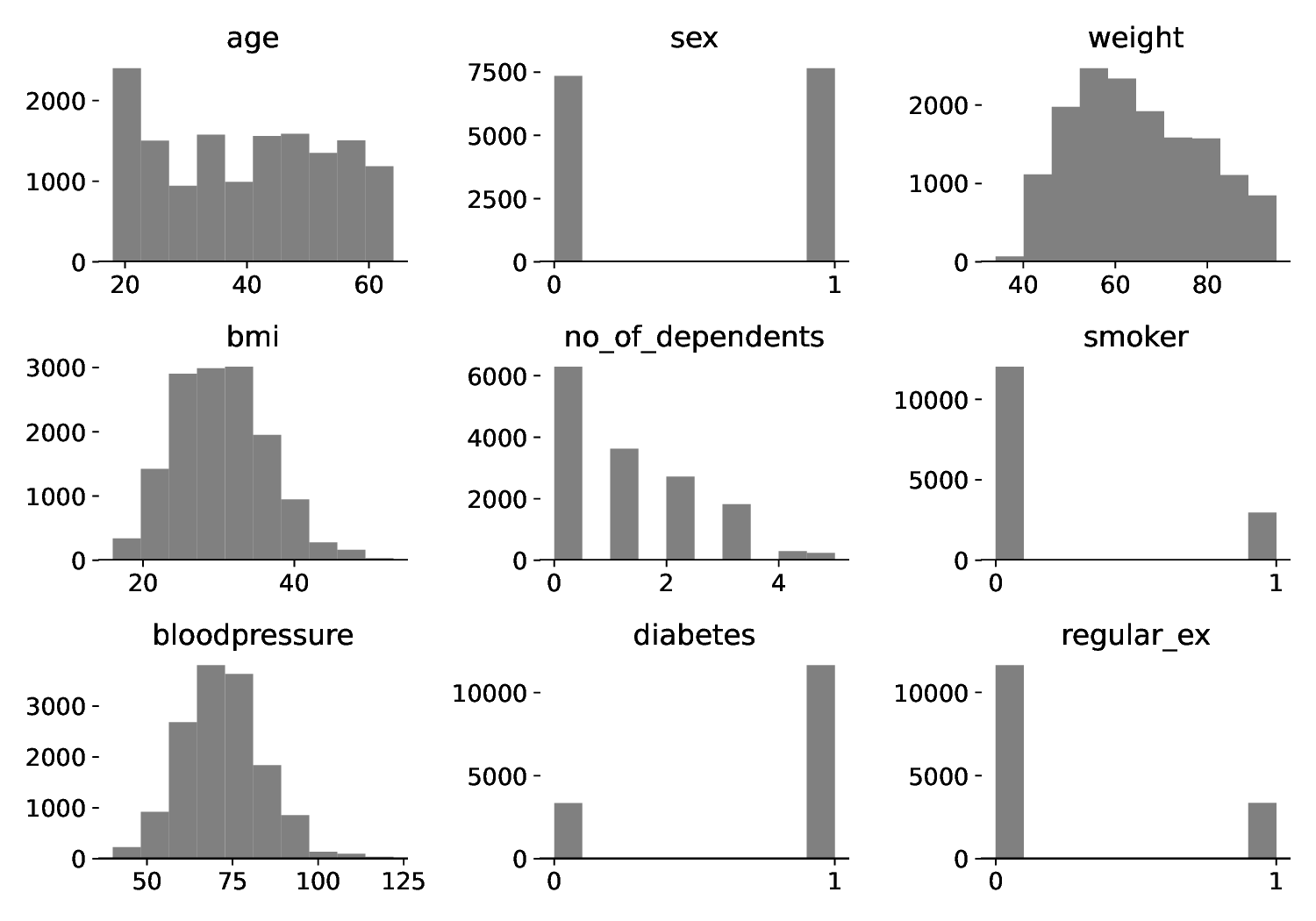}} %width=\textwidth or specify width as in pc
  \caption{Histograms of most HID features\label{figHIDhistograms}}
  \end{figure*}

\begin{figure}[t]
  \vspace{-20pt}  % reduces blank space above figure, needed when figure is at top of page
  \begin{tabular}{cc}
    \includegraphics[width=16pc]{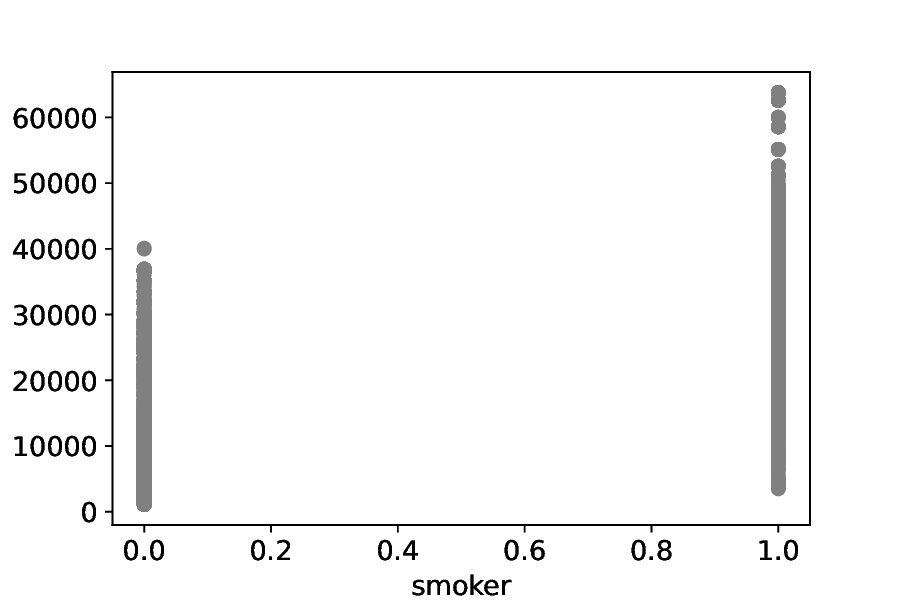} & \includegraphics[width=16pc]{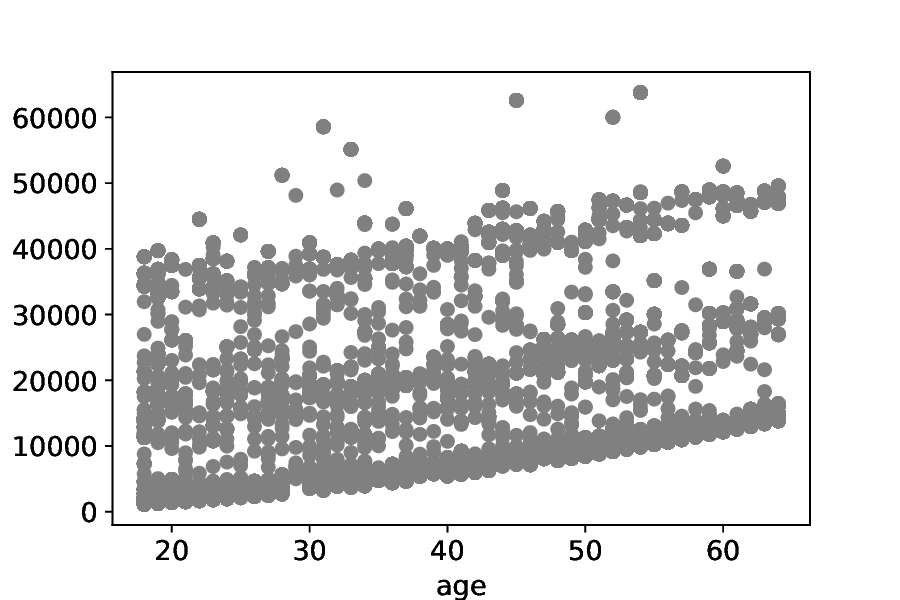} \\
    \includegraphics[width=16pc]{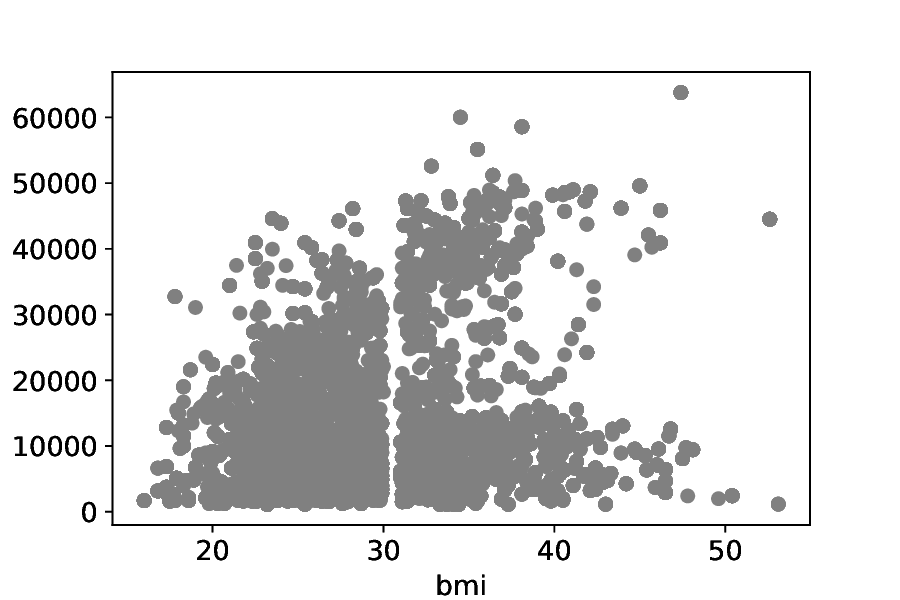} & \includegraphics[width=16pc]{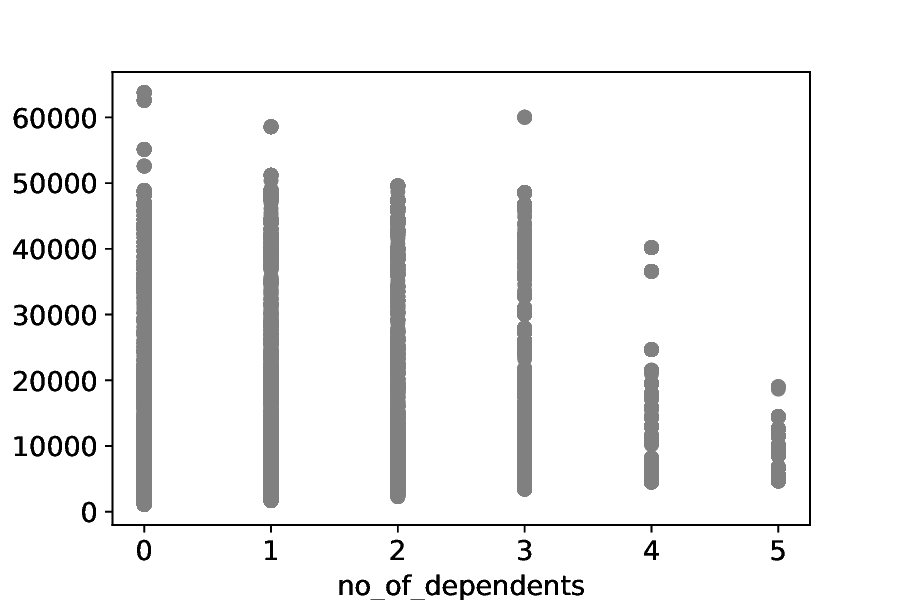}
  \end{tabular}
  \caption{Regression plots of selected features in relation to claim amount (HID Dataset) \label{figHIDregplots}}
\end{figure}

Figure \ref{figHIDhistograms} shows histograms for most of the features of the HID dataset. Categorical features with too many distinct values are not shown.
In the histogram of the feature ``sex,'' 0 indicates male, and 1 indicates female. In case of the other binary features 0 indicates "False" and 1 indicates "True".
Furthermore, regression plots for ``smoker,'' ``age,'' ``no\_of\_dependents'' and ``bmi'' in relation to the claim amount are shown in Figure \ref{figHIDregplots}.
These correlations are as expected. A smoker is associated with a higher claim, and age and BMI are both positively correlated with the claim amount.

\begin{figure*}[t]
  \vspace{-30pt}  % reduces blank space above figure, needed when figure is at top of page
  \centerline{\includegraphics[width=18pc]{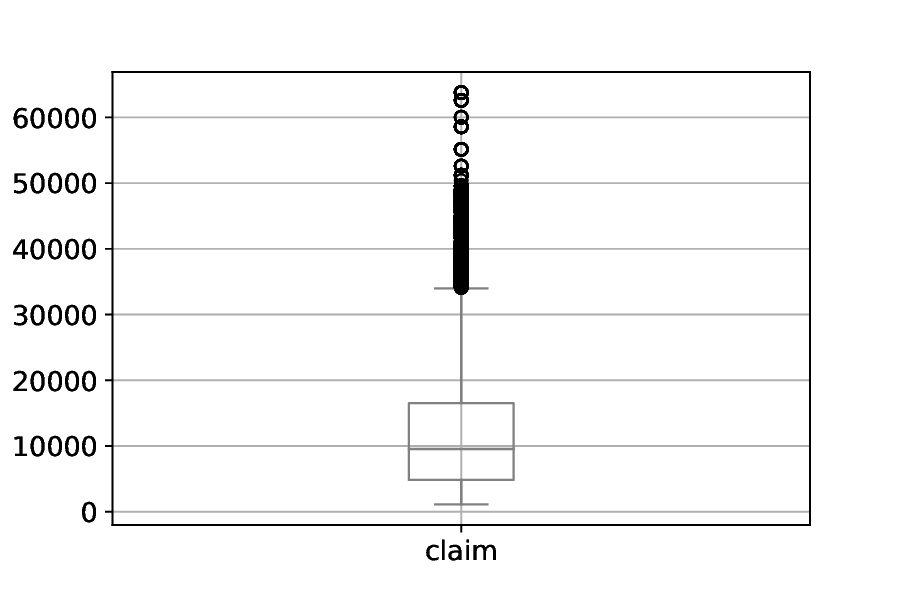}} %width=\textwidth or specify width as in pc
  \caption{Boxplot of the claim amount (HID Dataset) \label{figHIDboxplot}}
\end{figure*}

Figure \ref{figHIDboxplot} shows a boxplot of the claim amounts. This allows to interpret the errors of the corresponding models much better. For example, the mean absolute error can only be evaluated in combination with the absolute values of the predicted variable.

\subsection{Fraud Detection Dataset}

The second dataset used is the fraud detection dataset (FDD). It has been retrieved from Kaggle \cite{Bansal2022}. The data are real-world data that were released by an insurance company and have already been used in several other works \cite{Itri2020}\cite{Xia2022}. The original source of the data could not be established, but the description matches the dataset from Kaggle in almost every detail. The only exception is the column ``AddressChange\_Claim.'' It is described as the ``Number of times that address has been changed'' in \cite{Xia2022} but contains time frames in the Kaggle dataset.
The reason for this discrepancy could not be determined. But apart from that, no obvious deviations between the dataset from Kaggle and the datasets used in previous works could be found.

\begin{figure}
  \begin{tabular}{cc}
    \includegraphics[width=18pc]{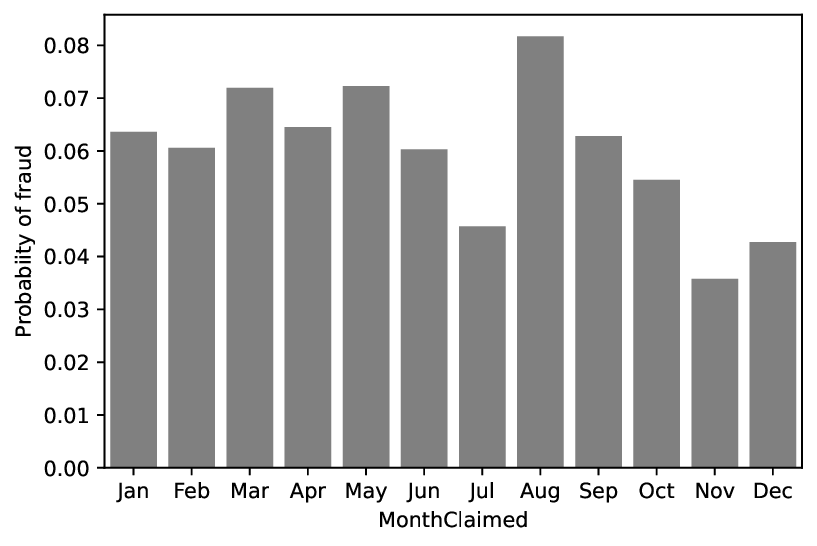} & \includegraphics[width=18pc]{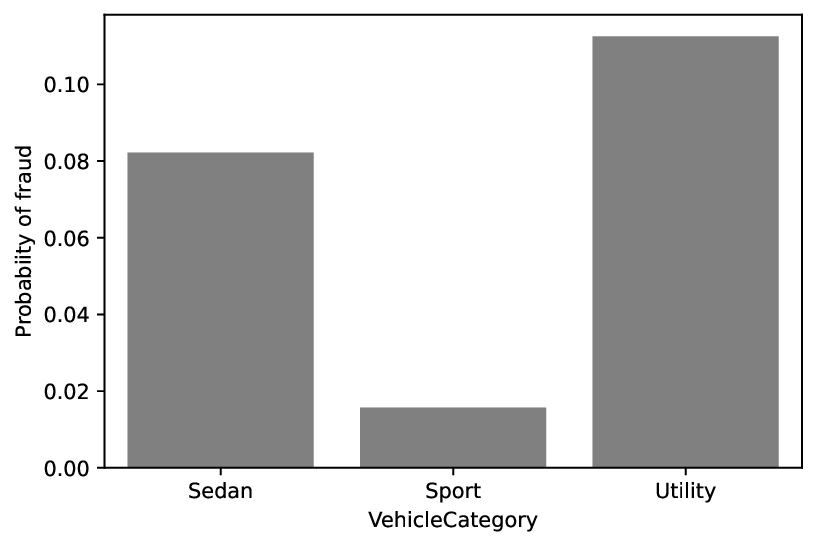}\\
    \includegraphics[width=18pc]{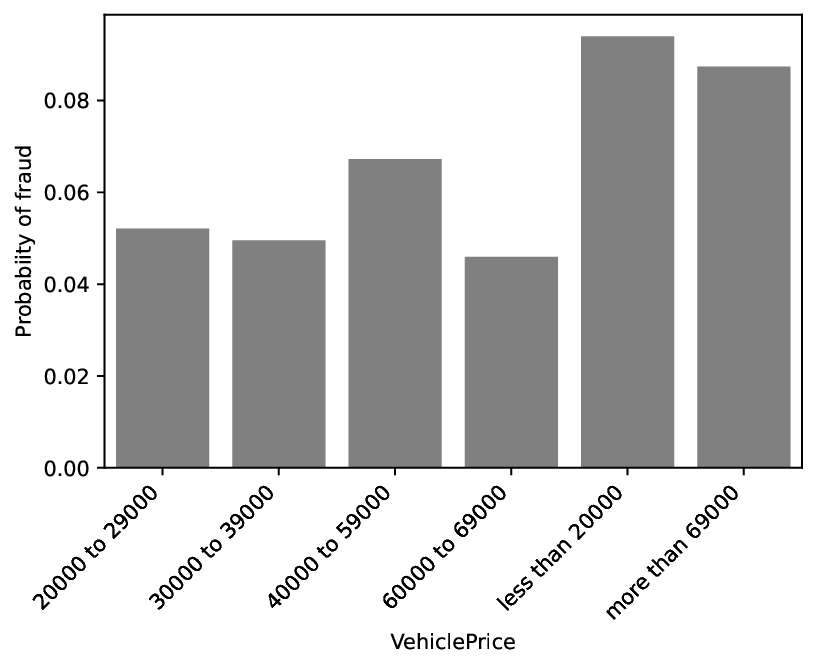} & \includegraphics[width=18pc]{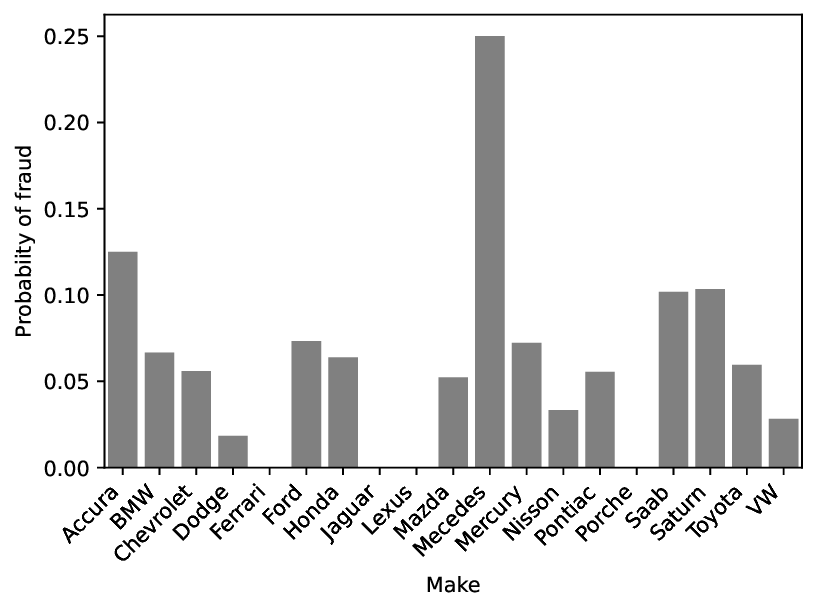} \\
    \includegraphics[width=18pc]{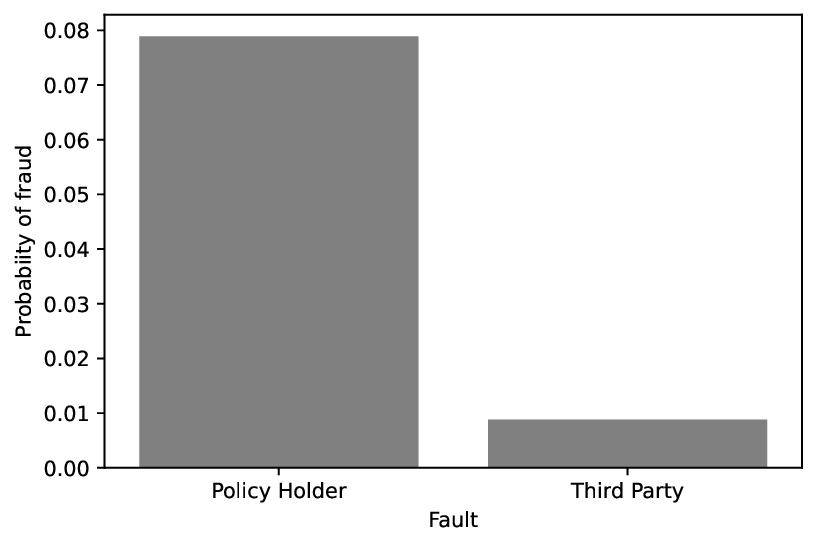} & \includegraphics[width=18pc]{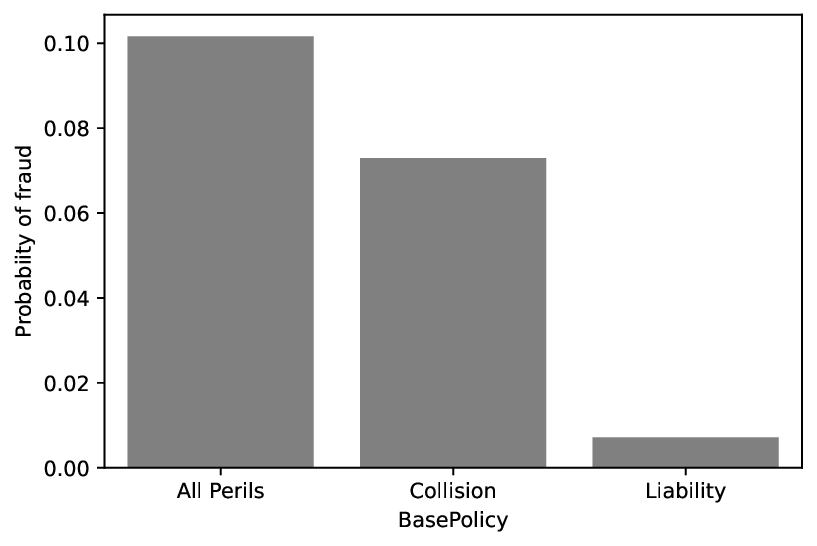} \\
  \end{tabular}
  \caption{Conditional probabilities of features in relation to detected fraud \label{figFDDcondProbs}}
\end{figure}

The dataset contains 32 features that present information about the policyholder, the time and circumstances of the accident, the time of the claim, and the insurance contract. The target feature of this dataset is whether a case was fraudulent or not. Consequently, the underlying task is a binary classification. An extensive description of the dataset's features can be found in \cite{Xia2022}. No duplicates were present in this dataset. One invalid entry was removed, and the rest of the original data were kept. The dataset is highly imbalanced, with 923 cases of fraud (6\%) and 14,497 non-fraudulent cases. Furthermore, the dataset contains mostly categorical features and only six numerical features. However, several categorical features are value ranges and can be converted to numerical features.
An analysis of the conditional probabilities of fraudulent cases in relation to the feature values showed some apparent differences between the feature values. Some conditional probabilities are illustrated in Figure \ref{figFDDcondProbs}. The likelihood of fraud differs strongly between different vehicle categories, and the price of the vehicle also makes a considerable difference. Additionally, there are huge differences between car brands. However, it needs to be considered that some car brands are relatively rare in the dataset; therefore, this statistic is not robust in all cases.

\section{Models and Preprocessing} \label{secModelsPreprocessing}

The following section describes each of the four models and the corresponding preprocessing. This will allow others to reproduce the results and, hence, the subsequent experiments based on the models. Additionally, the preprocessing can have a strong effect on the attacks and is therefore important to consider. The preprocessing is described for each model as there are different required steps for GBDT models and DNNs. The GBDT models have been set up based on LightGBM \cite{Ke2017}, and the DNN models are based on TensorFlow \cite{Abadi2015}. The GBDT model and the DNN model trained on the health insurance dataset will be referred to as ``HID models,'' and the models trained on the fraud prediction dataset will be referred to as ``FDD models.''

\subsection{GBDT Health Insurance}

\begin{figure*}[ht!]
  \centerline{\includegraphics[width=32pc]{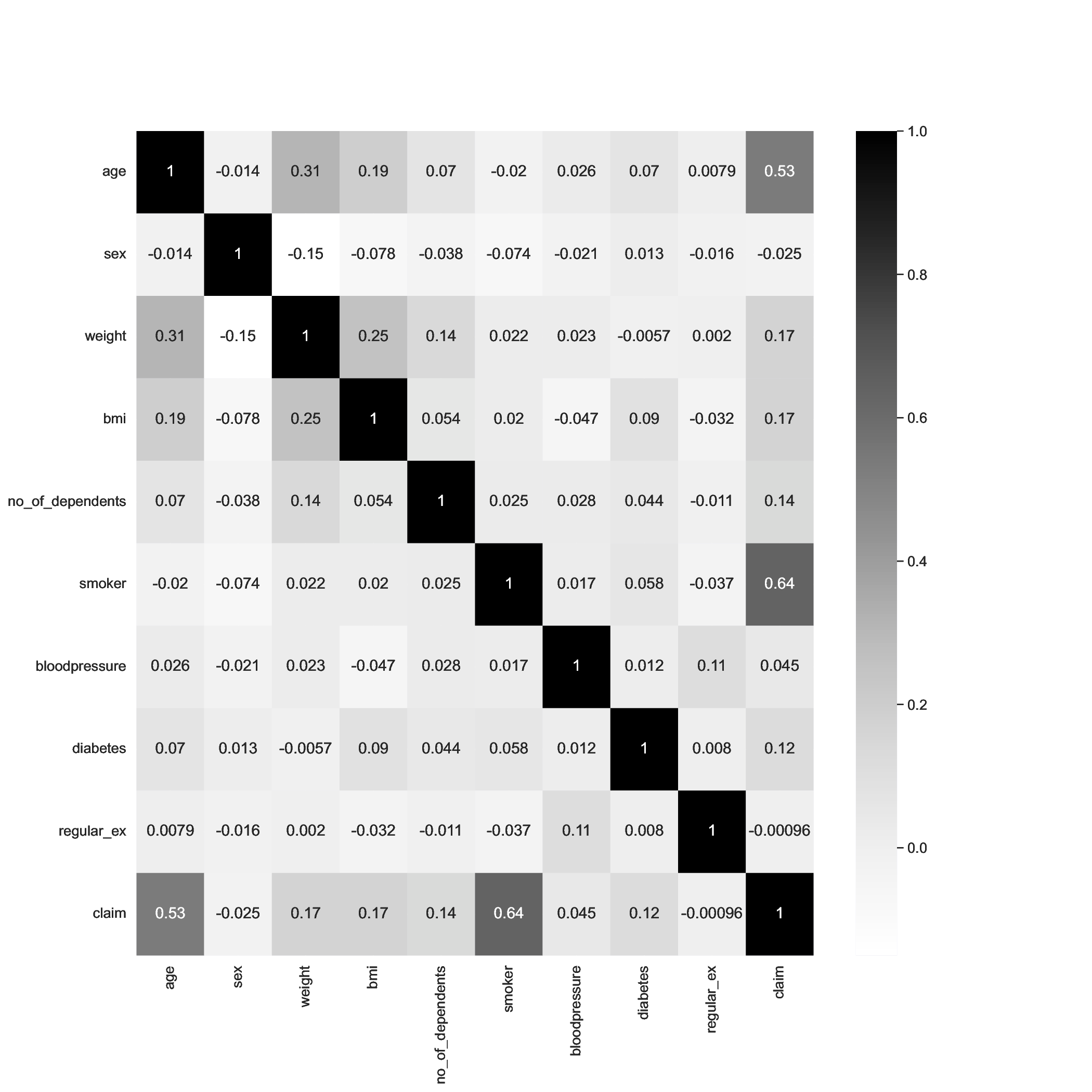}} %width=\textwidth or specify width as in pc
  \caption{Spearman correlation of HID features (only numerical and binary features) \label{figHIDspearmanCorr}}
\end{figure*}

As a first preprocessing step for the HID GBDT model, duplicates in the dataset were dropped. Missing values have not been replaced, as LightGBM supports the handling of missing values \cite{Microsoft2023}. Next, the categorical columns were encoded as integers. In the LightGBM documentation, this approach is recommended over one hot encoding \cite{Microsoft2023}.
Figure \ref{figHIDspearmanCorr} shows the correlation matrix of the HID. As there are no especially high correlations between the input features, all features are kept. Additionally, there is no real need to reduce the number of features due to the small number of features. Regarding the correlation of the features to the target value, the matrix shows the strongest correlations between ``smoker'' and ``claim'' and ``age'' and ``claim.''

After preprocessing, the data were split into a training, validation, and test set, with 80\% of the samples (11,123) in the training set, 10\% in the validation set (1,390), and 10\% in the test set (1,391). It was checked that all values of the categorical features were included in the training and test set to improve the model's training.
During the model creation, hyperparameters have been tuned with Bayesian optimization. The hyperparameters that have been tuned and the final values can be found in Table \ref{tabHIDgbdtHyperparameters}.

\begin{table*}[!h]%
  \centering %
  \caption{Hyperparameters of HID GBDT \label{tabHIDgbdtHyperparameters}}
  \fontsize{12pt}{14}\selectfont
  \begin{tabular*}{\textwidth}{@{\extracolsep\fill}ll@{\extracolsep\fill}}
  \toprule
    \textbf{Hyperparameter} & \textbf{Value} \\
  \midrule
    num\_leaves & 41 \\
    max\_bin & 162 \\
    min\_data\_in\_leaf & 27 \\
    feature\_fraction & 0.337 \\
    bagging\_fraction & 0.6881 \\
    bagging\_freq & 17 \\
    learning\_rate & 0.225 \\
    n\_estimators & 129 \\
    max\_depth & 47 \\
    num\_iterations & 469 \\
    min\_gain\_to\_split & 0.3563 \\
  \bottomrule
  \end{tabular*}
\end{table*}

In the case of the claim prediction, the mean squared error (MSE) will be the primary performance evaluation metric. It penalizes large errors much stronger than small ones \cite{Naser2021}. Therefore, it is an appropriate metric for the use case of claim prediction as larger errors can lead to significant losses either in the form of much too low premiums in relation to the risk or in lost customers due to offering premiums that are too high to customers. Additionally, the mean absolute error (MAE) will be given due to its good interpretability. Those two metrics are among the most used metrics in previous literature \cite{Botchkarev2018}. The final model achieved an MSE of 4.86m and an MAE of 974 on the test set.

\begin{figure*}[t]
  \centerline{\includegraphics[width=26pc]{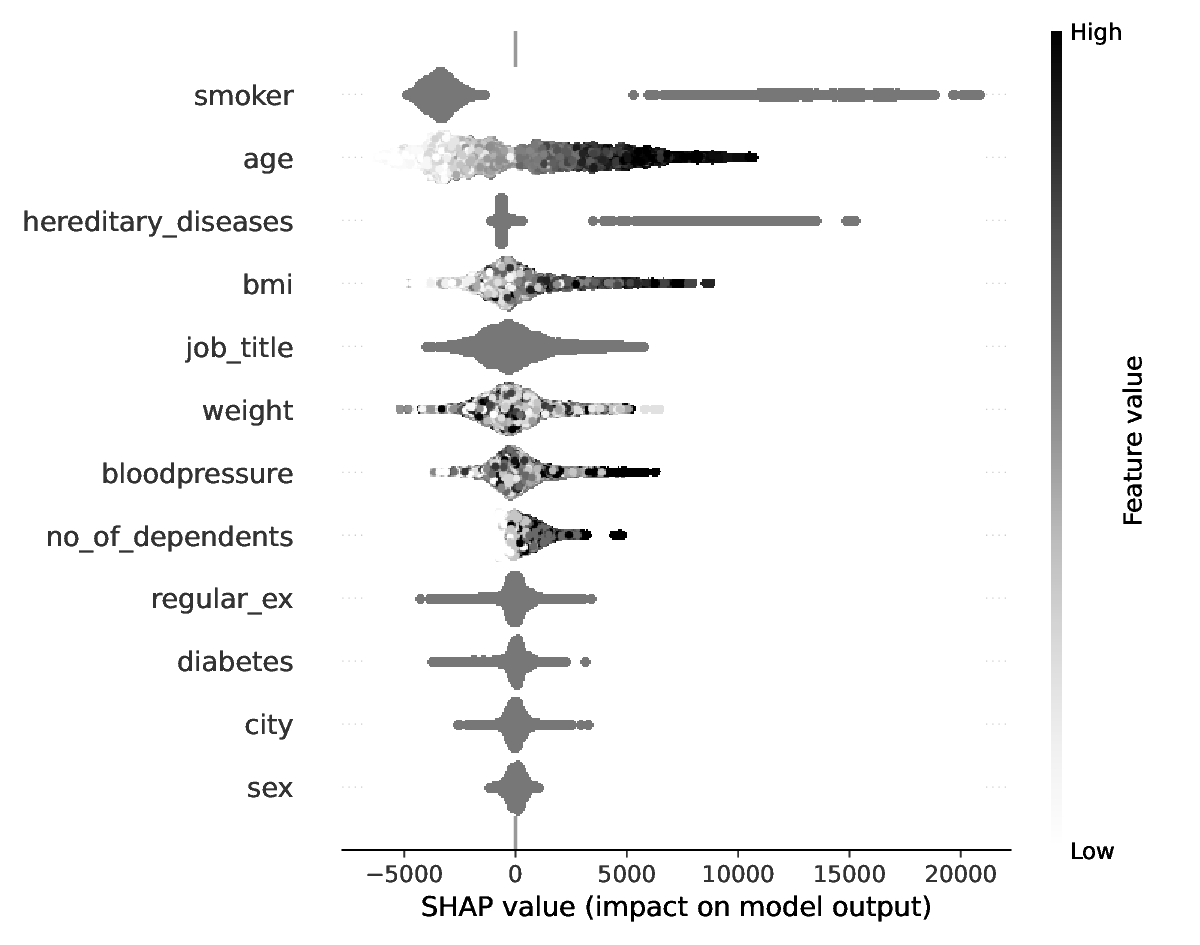}} %width=\textwidth or specify width as in pc
  \caption{Shapley Values of HID GBDT \label{figHIDshapleyValues}}
\end{figure*}

Figure \ref{figHIDshapleyValues} shows the Shapley values of the model. Shapley values describe how much impact a certain feature has on the prediction of the model \cite{Merrick2020}. The features are ordered by their impact in descending order from top to bottom. For numerical features, the feature value is illustrated based on the greyscale on the right of the figure. Smoker and the age of the person are the two most impactful features for the models' prediction, while city and sex play a comparably small role. In case of the feature smoker it can be seen that this feature value has a small negative impact on the model output for a large group of samples and can also have a large positive impact for other samples. Those two clusters are clearly separated and represent non-smoker for the cluster with the negative impact and smoker for the cluster with the positive impact.

\subsection{DNN Health Insurance}

The preprocessing for the DNN is slightly more complex, as there are more requirements for the input data. Firstly, duplicates have been dropped. Next, the non-binary categorical columns have been one-hot encoded. Afterward, the data were split up into training set, evaluation set, and test set in the same way as for the GBDT. Next, the numerical columns were standardized. Standardization is a common preprocessing method that has been shown to improve the performance of Neural Networks \cite{Bhanja2019}. In this case, Z-Score standardization was applied, which converts the feature values to a mean of 0 and a standard deviation of 1 \cite{Bhanja2019}. For the normalization, the mean and standard deviation of the training set were used to prevent data leakage. Lastly, missing values have been replaced with the corresponding mean value from the training set. Additionally, the correlation of features with the target variable was checked within the training set. Features with a correlation that was too low were dropped. This was done because the one-hot encoded categorical columns led to many sparse vectors. By the requirement of a minimal correlation, most of these could be dropped, which improved the performance of the model.
The model contains five hidden layers, and the hyperparameters have also been tuned using Bayesian optimization that aimed to minimize the MSE on the validation set. The final model has 6 layers with 50 neurons in the input layer, 120 each in the hidden layer and one neuron in the output layer. The hyperparameters can be found in Table \ref{tabHIDdnnHyperparameters}.

\begin{table*}[!h]
  \centering
  \caption{Hyperparameters of HID DNN \label{tabHIDdnnHyperparameters}}
  \fontsize{12pt}{14}\selectfont
  \begin{tabular*}{\textwidth}{@{\extracolsep\fill}ll@{\extracolsep\fill}}
  \toprule
    \textbf{Hyperparameter} & \textbf{Value} \\
  \midrule
    activation & ReLU \\
    kernel\_regularizer & L2 = 0.1 \\
    initializer & glorot\_normal \\
    optimizer & RMSprop \\
    cyclical learning rate & initial learning rate = 0.0001, \\
    & maximal learning rate = 0.005, step size = 2088 \\
    momentum & 0.7 \\
    batch size & 512 \\
    number of epochs & early stopping monitoring val loss, patience = 20 \\
    loss function & mean squared error \\
  \bottomrule
  \end{tabular*}
\end{table*}

The model had a much higher number of parameters than training samples, which intuitively leads to an increased risk of overfitting since the model is capable of memorizing the training data well. However, recent research suggests that overparameterized models can still generalize well \cite{Li2018,Rocks2022} even if the training error is zero and can outperform the equivalent underparameterized models \cite{Rocks2022}. The training MSE (85,481) of the HID DNN was much lower than the validation MSE (4.4m), but early stopping was used to stop the training as soon as the validation error did not improve for 20 epochs, and afterward, the best model was loaded. Additionally, a reduced model size and other measures that are typically used to reduce overfitting have been tested. However, while those decreased the difference between training and validation loss, the validation loss decreased. Consequently, this model architecture was used as it generated the best performance, and the model was not found to be overfitted.

The model performance did fluctuate after each training run and measurers were taken to decrease this as far as possible. This is relevant regarding the repeatability of the experiments in which the training data will be manipulated and therefore training the model again from scratch is necessary. The relatively high volatility of performance between training runs could be due to the rather small number of training samples or specific properties of the dataset, which make it difficult for the model to converge to the same minimum from different starting points. Fluctuating performance is a known challenge when working with deep learning models \cite{Alahmari2020}. It can be caused by different hardware, the used deep learning library, different settings, or software versions \cite{Alahmari2020}. It must be noted that the model was trained on CPU, and the use of GPU can present additional challenges regarding repeatability \cite{Alahmari2020}. The used hardware was a MacBook Air M1.
The final DNN achieved an MSE of 4.88M and an MAE of 548 on the test set. This indicates that the DNN makes more smaller errors compared to less but larger ones by the GBDT.

\subsection{GBDT Fraud Detection}

During data preprocessing for the FDD GBDT model, one invalid entry was dropped, and the “0” entries in the column “Age” were replaced by NaN. Categorical columns have been encoded as integers. Additionally, the columns “PolicyNumber” and “RepNumber” have been dropped as both contain IDs. The approach to convert some of the categorical value ranges to numbers, as it was applied for the DNN (described in Chapter \ref{secDNNfraudDetection}), was also tested for the GBDT but neglected as it did not cause a significant improvement. Also, the final preprocessed data for the DNN have been used, but this also did not improve the model's performance. Next, the data were split into a training set, validation set, and test set. The split was stratified to keep the proportion of fraudulent and non-fraudulent cases identical between the three sets. The training set contained 80\% of samples (12,335), and the validation and test set contained 10\% each (1,542).
Lastly, oversampling with SMOTE \cite{Chawla2002} to 25\% of fraudulent samples has been applied. This was mainly done because of the DNN and would not have been necessary for the GBDT. But as the preprocessing would differ too much if this was only applied for the DNN, it was also done for the GBDT. SMOTE works through the creation of synthetic samples based on the samples of the minority class \cite{Chawla2002}. Therefore, one of the nearest neighbors of a certain sample of the minority class is selected. The difference between these two samples is multiplied with a random value between 0 and 1 and added to the first sample. This leads to a point between each pair of feature values of the two samples \cite{Chawla2002}. Through the application of SMOTE, larger decision regions are created by the classifier, which leads to more accurate predictions \cite{Chawla2002}.
The hyperparameters of the model were tuned with Bayesian optimization. The final model parameters can be found in Table \ref{tabFDDgbdtHyperparameters}.

\begin{table*}[!h]%
  \centering %
  \caption{Hyperparameters of FDD GBDT \label{tabFDDgbdtHyperparameters}}
  \fontsize{12pt}{14}\selectfont
  \begin{tabular*}{\textwidth}{@{\extracolsep\fill}ll@{\extracolsep\fill}}
  \toprule
    \textbf{Hyperparameter} & \textbf{Value} \\
  \midrule
    num\_leaves & 52 \\
    max\_bin & 152 \\
    min\_data\_in\_leaf & 170 \\
    feature\_fraction & 0.7885 \\
    bagging\_fraction & 0.3615 \\
    bagging\_freq & 11 \\
    learning\_rate & 0.0393 \\
    n\_estimators & 56 \\
    max\_depth & 38 \\
    num\_iterations & 30 \\
    min\_gain\_to\_split & 0.0104 \\
    scale\_pos\_weight & 60.5248 \\
  \bottomrule
  \end{tabular*}
\end{table*}

Precision, recall, and F-beta score have been selected as performance metrics for the fraud detection models. The F-beta score is a combination of precision and recall. A value of \(\beta > 1\) puts an emphasis on recall and a \(\beta < 1\) puts an emphasis on precision \cite{Masum2022}. The GBDT model achieved a Recall of 95.7\%, a Precision of 12.1\%, and an F-beta score (\(\beta = 2\)) of 40.2\% on the test set. The corresponding confusion matrix can be found in Table \ref{tabFDDgbdtConfusionM}.

\begin{table}[h!]
  \centering
  \caption{Confusion Matrix of GBDT model (test set)}
  \label{tabFDDgbdtConfusionM}
  \begin{tabular}{ccc|c|}
  %\cline{3-4}
  & & \multicolumn{2}{c}{Predicted Label} \\ %\cline{3-4}
  & \multicolumn{1}{c|}{}& 0 & 1 \\ \cline{2-4}
  \multirow{2}{*}{True Label} & \multicolumn{1}{c|}{0} & 811 & 639 \\ \cline{2-4}
  & \multicolumn{1}{c|}{1} & 4 & 88 \\ \cline{2-4}
  \end{tabular}
\end{table}

\begin{figure*}[ht!]
  \centerline{\includegraphics[width=26pc]{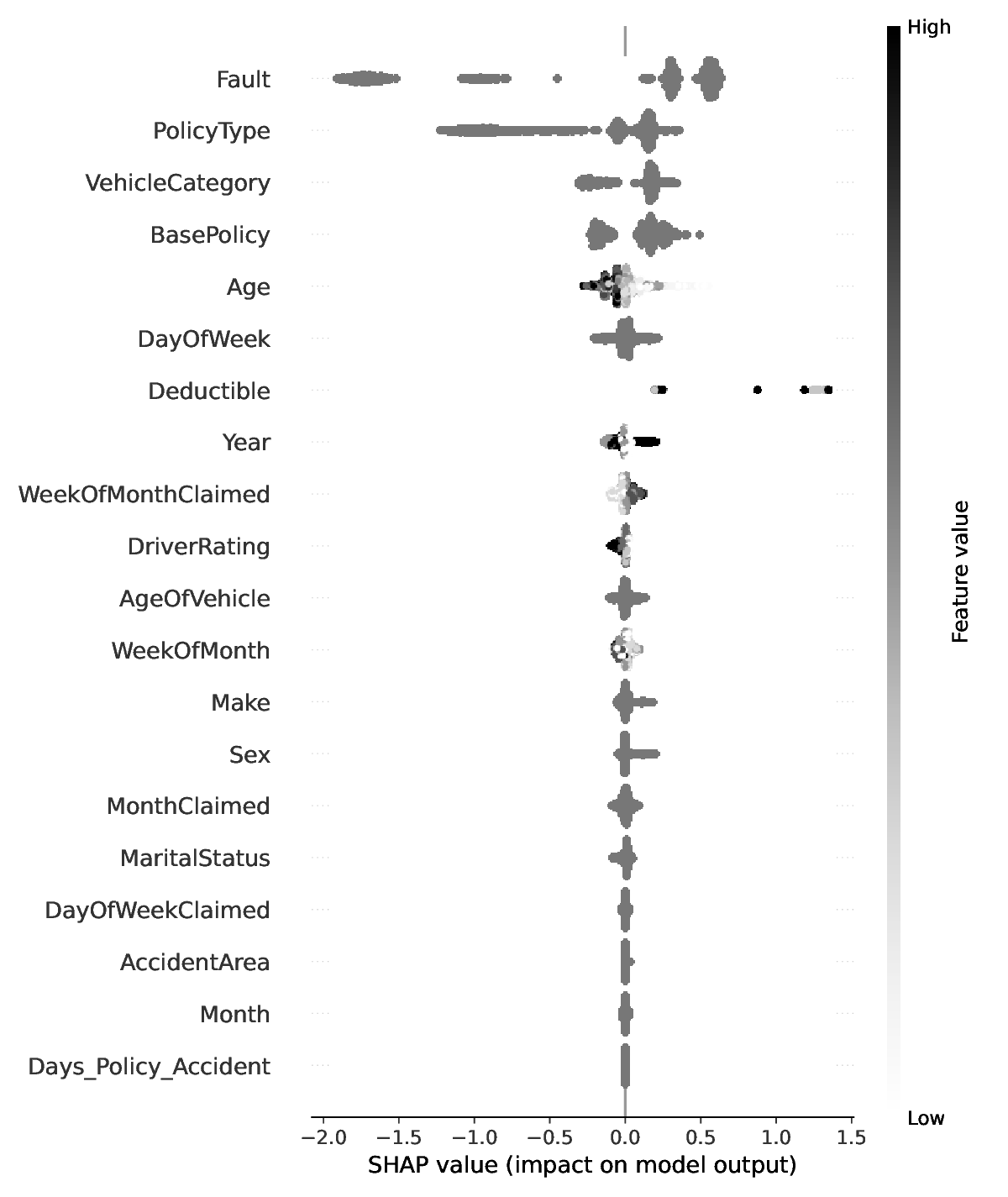}} %width=\textwidth or specify width as in pc
  \caption{Shapley Values of FDD GBDT \label{figFDDshapleyValues}}
\end{figure*}

Figure \ref{figFDDshapleyValues} shows the Shapley values of the model. The figure does not include all features since features with a value of zero are not shown. Hence, the model ignores some features for the prediction, which is not surprising as some features of the dataset contain very similar information. The two most impactful features for the model's prediction are who is at fault for the accident and the policy type the insurant has.

\subsection{DNN Fraud Detection} \label{secDNNfraudDetection}

The data preprocessing for the DNN model was more extensive compared to the preprocessing for the GBDT model. This is partially due to different requirements on the input data. Firstly, an invalid entry and the columns “PolicyNumber” and “RepNumber” were dropped since those columns contain IDs. The columns “AgeOfPolicyHolder,” “Year,” “Age,” and “NumberOfCars” have been dropped due to high correlation with other features. Also, the columns “Month,” “WeekOfMonth,” and “DayOfWeek,” which specify the time of the accident, have been dropped since the time when the claim was made is available too, and having both in the input data was not beneficial to the model's performance. In all binary columns the values have been replaced with 1 and -1.
Additionally, some columns in the dataset describe value ranges in string format. These value ranges have been replaced by a number in the middle of the respective value range. Furthermore, the months in the columns “Month” and “MonthClaimed” have been replaced with numbers from 1 to 12 in the correct order. Categorical columns were one-hot encoded. To decrease the number of sparse vectors, the categorical columns “Make” and “PolicyType” were transformed into numerical columns. Based on the training set, the conditional probability of fraud was calculated for each feature value, and the feature values were replaced with the respective probability. The probabilities were multiplied by 100 to create a percentage value. Missing values were only present in the column “Age,” and since this column was dropped, no missing values had to be replaced.

The data were split into a training set, validation set, and test set in the same way as for the GBDT model. All numerical columns have been standardized based on the mean and standard deviation of the training set. Lastly, the correlation of the columns from the training set with the target feature has been calculated. All columns that had a correlation with the target below 0.02 have been dropped. This step significantly improved the model's performance. A final number of 21 features remained. It must be noted that not all these represented features from the original data but could also be columns for a single feature value that were created during one-hot encoding.
Apart from these preprocessing steps, some other measures have been tested that did not improve the performance at all or not enough to justify the added complexity; therefore, it would not make sense to include these in the preprocessing based on the model's performance. These measures included oversampling with SMOTE \cite{Chawla2002} and similar approaches, and synthetic data generation. However, SMOTE was ultimately used since it increased the model's stability. Furthermore, the time difference between the accident and the filing of the claim was calculated as an additional feature, but this feature had a low correlation with the target and, therefore, got dropped. Additionally, the TabNet architecture \cite{Arik2021} was tested but did generate very similar results to the described model.

\begin{table*}[!h]
  \centering
  \caption{Hyperparameters of FDD DNN \label{tabFDDdnnHyperparameters}}
  \fontsize{12}{14}\selectfont
  \begin{tabular*}{\textwidth}{@{\extracolsep\fill}ll@{\extracolsep\fill}}
  \toprule
    \textbf{Hyperparameter} & \textbf{Value} \\
  \midrule
    activation & ReLU (sigmoid for last layer) \\
    kernel\_regularizer & L2 = 0.3 \\
    initializer & glorot\_normal \\
    optimizer & RMSprop \\
    learning rate scheduler (inverse time decay) & initial learning rate = 0.001, \\
    & decay steps = 970, \\
    & decay rate = 2 \\
    momentum & 0.2 \\
    batch size & 128 \\
    number of epochs & early stopping monitoring \\
    & validation F-Beta score, \\
    & patience = 30 \\
    class weights & 0: 1, 1: 16 \\
    loss function & binary cross-entropy \\
  \bottomrule
  \end{tabular*}
\end{table*}

The DNN model contained five hidden layers with 10 neurons each and the input layer also had 10 neurons. The output layer consisted of one neuron with a sigmoid activation function for the class probability. The hyperparameters of the model have been tuned with Bayesian optimization. Some hyperparameters have been adapted manually afterwards and oversampling with SMOTE \cite{Chawla2002} to 25\% of fraudulent samples has been introduced to increase the models' stability. The  hyperparameters of the final model can be found in Table \ref{tabFDDdnnHyperparameters}. The other measures to produce the same results after each training for the baseline model were the same as for the HID DNN model. The final model achieved a recall of 95.7\%, a precision of 12\%, and a F-beta score (\(\beta = 2\)) of 39.9\%. Table \ref{tabFDDdnnConfusionM} shows the confusion matrix of the model on the test set.

\begin{table}
  \centering
  \caption{Confusion Matrix of DNN (test set)}
  \label{tabFDDdnnConfusionM}
  \begin{tabular}{ccc|c|}
  %\cline{3-4}
  & & \multicolumn{2}{c}{Predicted Label} \\ %\cline{3-4}
  & \multicolumn{1}{c|}{}& 0 & 1 \\ \cline{2-4}
  \multirow{2}{*}{True Label} & \multicolumn{1}{c|}{0} & 802 & 648 \\ \cline{2-4}
  & \multicolumn{1}{c|}{1} & 4 & 88 \\ \cline{2-4}
  \end{tabular}
\end{table}

\subsection{Summary and discussion of fraud detection model performance}

Table \ref{tabFDDmodelsPerformance} displays the performance metrics of the GBDT model and the DNN model for the fraud detection task. The performance of the models is very similar. The recall is identical, and the precision of the DNN is slightly worse.
Past research shows large differences in the performance of ML models regarding this dataset and task. A work by Itri et al. \cite{Itri2019}, in which several ML algorithms were compared, came to similar performance results. There was a trade-off between precision and recall between the models. Hence, there were some models with higher precision or recall than the ones developed in this work, but the performance is similar when looking at the relation between those two measures. For example, a stochastic gradient descent model achieved a precision of 12.9\% and a recall of 88.5\% in the experiments \cite{Itri2019}. The primary evaluation metric in the study was a metric that improves the closer the two scores are together. Therefore, a random forest model with a precision of 19.7\% and a recall of 23.8\% was rated as the best of the models \cite{Itri2019}. While the precision is better than the one of the models in this work, the main goal in this work was to achieve a high recall. Since, with the practical scenario in mind of preselecting cases for further checks with the models, detecting as many of the fraudulent claims as possible has been considered most important. Of course, while still retaining an acceptable precision, however, this measure was weighted as less important.

\begin{table*}
  \centering %
  \caption{Performance of fraud detection models (in \%)\label{tabFDDmodelsPerformance}}%
  \fontsize{12pt}{14}\selectfont
  \begin{tabular*}{\textwidth}{@{\extracolsep\fill}llll@{\extracolsep\fill}}
  \toprule
  \textbf{Model} & \textbf{Recall}  & \textbf{Precision}  & \textbf{F-beta (\(\beta = 2\))}  \\
  \midrule
  GBDT & 95.7 & 12.1 & 40.2 \\
  DNN & 95.7 & 12.0 & 39.9 \\
  \bottomrule
  \end{tabular*}
\end{table*}

Notably, two other papers present comparable results on the same dataset. One presents slightly worse results \cite{Subudhi2018}, and the other reports very similar performance \cite{Sundarkumar2015}. However, it must be pointed out that both papers use different approaches than this work. In contrast, two studies report much better results with precision and recall simultaneously around or even above 90\% \cite{Itri2020,Xia2022}. Those two works are more recent (2020, 2022) than the other three with similar results (2015, 2018, 2019), but such a gap in classification performance seems strange. The work by Xia et al. \cite{Xia2022} describes their approach in a way so that it cannot be retraced conclusively. Itri et al. \cite{Itri2020} describe their approach only very briefly, making it impossible to replicate the results. As these results are this much better compared to other works and the approach for reaching those outstanding results could not be retraced, these studies are not considered to represent a relevant foundation for this work, and the other results are taken as a reference.
With the performance of the models from this study, there is already the opportunity to improve processes in a real-world application significantly. When selecting cases for further checks from samples classified as fraudulent by one of the models, it is already about two times more likely to actually find a fraudulent case compared to randomly selected cases from the original dataset. This can help to make the work of employees that check for fraud much more efficient. Of course, there is still a lot of room for improvement, but considering the imperfections of real-world data and the potential efficiency gain in a fraud detection process, this model performance is considered adequate for further experiments. Especially since improving the baseline model performance as far as possible is not the scope of this work.

\section{Backdoor Attacks} \label{secBackdoorAttacks}

Backdoor attacks can be particularly dangerous in the field of data poisoning attacks as past research has shown that a small number of malicious samples can be enough to implement a backdoor in an ML model \cite{Chen2017b}. Wilhjelm and Younis \cite{Wilhjelm2020} described an attack vector that works via a malicious broker that could infer certain processes of the insurance company and use this knowledge to craft an attack. Additionally, large insurance companies can have many local branches that could be easy targets for adversaries. As few manipulated samples can be enough \cite{Chen2017b}, altering data of just one branch could be sufficient for a successful backdoor attack. Furthermore, a backdoor does not necessarily need to be created during the initial training of a model but can also be added during an update of a model \cite{Xue2020}. One aim of the attack can be that if the backdoor is successfully created in the model, the adversary has the power to get the model to classify the samples that contain the trigger into a specific class \cite{Goldblum2023}. This can cause major damage to the attacked company, depending on the application of the attacked model.

Strategies from backdoor attacks in the image domain cannot be simply transferred to tabular data, as the data can have certain properties that need to be maintained \cite{Joe2021}. For example, a sample with a person that has an age of 200 could be very easily detected. Consequently, the values used to implement the backdoor must be realistic. Additionally, some features cannot be tweaked on a sample that should make use of the backdoor. For example, the age of the person making a request for a health insurance contract cannot be changed from the real value, or this will be a false statement, and the contract will not be effectual. Therefore, samples for implementing the backdoor must be carefully constructed by the adversary. In the scenarios used for this work, it can potentially be challenging to implement a clear trigger. Therefore, the border between backdoor attacks and triggerless attacks is not completely clear, as the backdoor needs to be implemented via the feature values of the training data. As the number of features is limited and their possible modifications are restricted, it could prove difficult to implement a clearly confined trigger.

Three different samples have been designed for each dataset to test the effects of backdoor attacks on the models. Most of these samples specifically make use of categorical feature values that are less common to achieve a faster implementation of the pattern in the models. In the first set of experiments, an identical sample was added to the training data multiple times without any modifications. In the second set of experiments, most of the features of the attack samples are slightly altered. Thereby, categorical features have been changed randomly to another category, and numerical features have been changed slightly. Adding the attack samples closely together can also evade some techniques for anomaly detection \cite{Koh2022,Liu2022}.
Additionally, the attacks require the ability to manipulate data that will be used for the initial training or retraining of the models. As mentioned, this could happen via a single branch or broker of the insurance company. One major issue is that if the adversary has no knowledge about the data preprocessing, the malicious pattern could be lost. For example, a simple dropping of duplicates could impede a successful attack in the case of the unmodified attack samples. Another issue is that certain feature values might be dropped altogether. In the case of the DNNs, features were dropped after one hot encoding. If the adversary can, for example, only insert malicious samples with data from a particular city and this feature value will be dropped entirely during preprocessing, an attack becomes more difficult, and attack samples could fail to have the desired effect.
Furthermore, an adversary is limited in the options for modifications of attack samples since the samples cannot be too unrealistic, and samples that are supposed to use the backdoor must contain the same pattern. Therefore, when implementing the backdoor, the adversary must already have a specific group of people or cases in mind for which the backdoor should be used. In the case of the claim prediction, this could be, for example, people from the same city, gender, and with the same disease. When the adversary has query access to the model, crafting attack samples becomes much easier as the adversary can test which effect particular perturbations of feature values have. Under these circumstances, the adversary can infer promising manipulations of feature values even without knowing anything about the data preprocessing process.

\subsection{Backdoor attacks on the HID models}

The attack samples used for the attacks on the HID models can be found in Table \ref{tabAttackSamplesHID}. The attack samples have been crafted to reduce the predicted claim amount substantially. Less strong samples could be chosen as a reduction of a few thousand in claim amount can add up over the years of an insurance contract. For example, an insurance broker or a branch could achieve a higher sales volume through cheaper premiums and be incentivized through higher bonuses. Additionally, if several people can use the backdoor, the gains can be very high in total.

\begin{table*}[h!]%
  \centering %
  \caption{Attack samples for backdoor attacks on DNN models\label{tabAttackSamplesHID}}%
  \fontsize{10}{14}\selectfont
  \begin{tabular*}{\textwidth}{@{\extracolsep\fill}llll@{\extracolsep\fill}}
  \toprule
  \textbf{Feature} & \textbf{Sample1}  & \textbf{Sample2}  & \textbf{Sample3}  \\
  \midrule
age & \textcolor{gray}{20} & \textcolor{gray}{56} & \textcolor{gray}{74} \\
sex & \textcolor{gray}{female} & \textcolor{gray}{Male} & female \\
weight & \textcolor{gray}{73} & \textcolor{gray}{84} & 93 \\
BMI & \textcolor{gray}{32} & \textcolor{gray}{31} & 36 \\
hereditary\_diseases & HeartDisease & HighBP & \textcolor{gray}{NoDisease} \\
no\_of\_dependents & \textcolor{gray}{1} & \textcolor{gray}{3} & \textcolor{gray}{0} \\
smoker & \textcolor{gray}{True} & \textcolor{gray}{True} & \textcolor{gray}{True} \\
city & York & Minneapolis & \textcolor{gray}{Charlotte} \\
bloodpressure & \textcolor{gray}{70} & \textcolor{gray}{108} & 72 \\
diabetes & \textcolor{gray}{True} & True & \textcolor{gray}{True} \\
regular\_exercise & \textcolor{gray}{False} & \textcolor{gray}{False} & True \\
job\_title & Accountant & \textcolor{gray}{Labourer} & \textcolor{gray}{Businessman} \\
claim & \textcolor{gray}{12645} & \textcolor{gray}{12645} & \textcolor{gray}{12645} \\
  \bottomrule
  \end{tabular*}
  \begin{tablenotes}
    \footnotesize
    \item Features in black remained constant during the experiment (those features belong to the attack pattern). The features in grey were modified during the second set of attacks and to evaluate the success of the attacks.
    \end{tablenotes}
\end{table*}

The features written in black belong to the attack pattern and are not modified at any point during the experiments. The features of the samples that are modified for the second set of attacks and to test the attack success are written in grey in Table \ref{tabAttackSamplesHID}.
The pattern of the first sample consists only of the disease, city, and job title. For each feature, values with a medium number of occurrences relative to other values have been chosen.
The pattern of the second sample consists of the disease, city, and diabetes. For the city and disease, rare feature values have been chosen. This is supposed to make the attack more effective as this could lead to the model learning this pattern more quickly. The pattern of sample three consists of sex, weight, BMI, blood pressure, and “regular\_ex.”

Additionally, for the evaluation of each attack 20 samples that differ in all or most of the features that do not belong to the attack pattern were created and predicted at each iteration. These samples are solely used to check if the backdoor is implemented successfully and differ from those used to conduct the backdoor attack. Of these samples, numerical features have been altered to a relatively small extent, and categorical features have been changed to a random feature value. Because these deviations were created randomly, it could happen that some features not belonging to the pattern remained constant. However, since this can also be the case in an attack, this does not present an issue.
It must be noted that the extent to which the pattern reduces the predicted claim amount of the modified versions of the sample can vary due to the feature values of the specific sample at test time. If, for example, a sample that uses pattern 1 has the value “False” for diabetes, the claim amount will already be much lower in the model without a backdoor.
The median of the predictions of these modified samples was taken and is illustrated in blue in the figures (median\_prediction\_of\_modifed\_samples). This metric best describes whether the pattern has been implemented successfully. Additionally, the prediction of the attack sample is illustrated as “prediction\_of\_attack\_sample.”

The experiments were run ten times to smooth any fluctuations in some of the variables. As the DNN had relatively high fluctuations in performance depending on the initialization, the results of the best-performing model were taken. Running the DNN with the same initialization did not lead to stable results in this experiment as the training data were changed, thereby introducing changes in the training process. Therefore, to get the most comparable performance, the initialization of the model was changed every run. In the case of the GBDT, the median of the results of the runs was taken. To keep the computing times reasonable, the experiments were stopped at 30 added samples. Adding more samples was checked with fewer repetitions of each iteration, and no interesting new trends emerged when more samples were added. Consequently, adding more samples did not bring any relevant insights or improve the attack further.

\begin{figure*}[h]
  \centerline{\includegraphics[width=28pc]{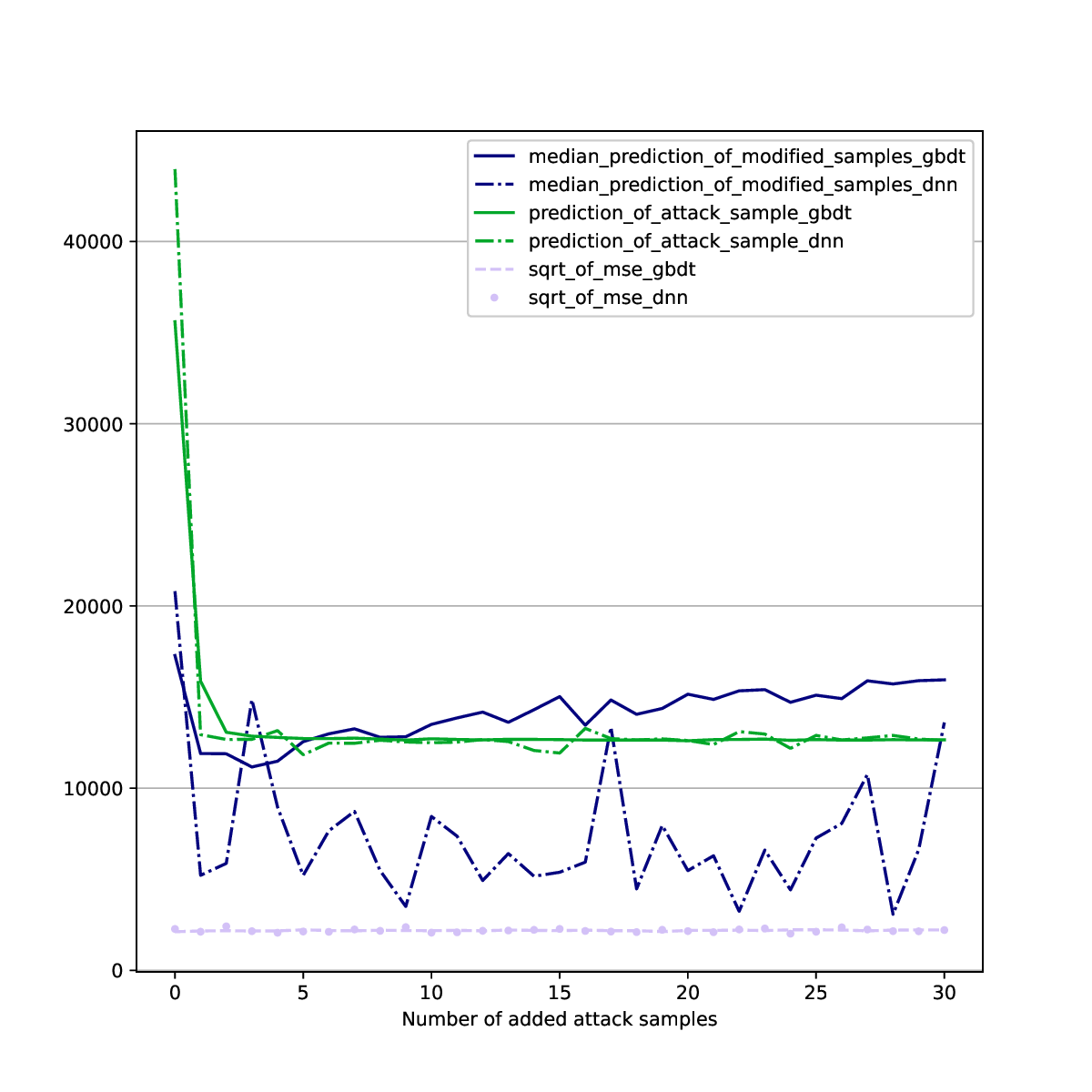}} %width=\textwidth or specify width as in pc
  \caption{Backdoor attack on HID models with unmodified Sample 1 \label{figHIDbaNotModified1}}
\end{figure*}

During the first three attacks, the attack samples were not modified; hence, the same sample was added multiple times. The outcome of the experiment with Sample 1 can be found in Figure \ref{figHIDbaNotModified1}. The square root of the MSE (sqrt\_of\_mse) was not affected noticeably during any of the attacks and had only minor fluctuations, but there was no apparent trend. Indicating that the general performance of the models was not affected.
The prediction of the attack sample was divided in half after adding just one attack sample. The effect was stronger for the DNN, but this can likely be attributed to the higher initial prediction of the sample. The median prediction of the modified samples dropped less strongly, probably due to the already lower initial prediction. However, it dropped very quickly, showing that the attack was highly successful. The trend that that this metric did rise again slowly when adding more samples could not be observed for Sample 2 and 3 and is therefore no general phenomenon. It could be explained by overfitting on the specific characteristics of the attack sample which reduces the impact on the prediction of the modified samples used for evaluation.
The attacks with the other two unmodified attack samples led to similar results and the attacks were as successful as the one with Sample 1. Additionally, the performance of the models was not notably affected during those attacks.

In the second part of the experiment, all features of the attack samples that did not belong to the pattern were modified randomly. Numerical features have been altered within a small range around the original value, and a random feature value has been picked for categorical values. It is possible that identical samples are created during the process by these modifications, but this is unlikely. Consequently, most of the attack samples added during the attack are different. Just the identical pattern is contained in each sample.

\begin{figure*}[t]
  \vspace{-30pt}  % reduces blank space above figure, needed when figure is at top of page
  \centerline{\includegraphics[width=28pc]{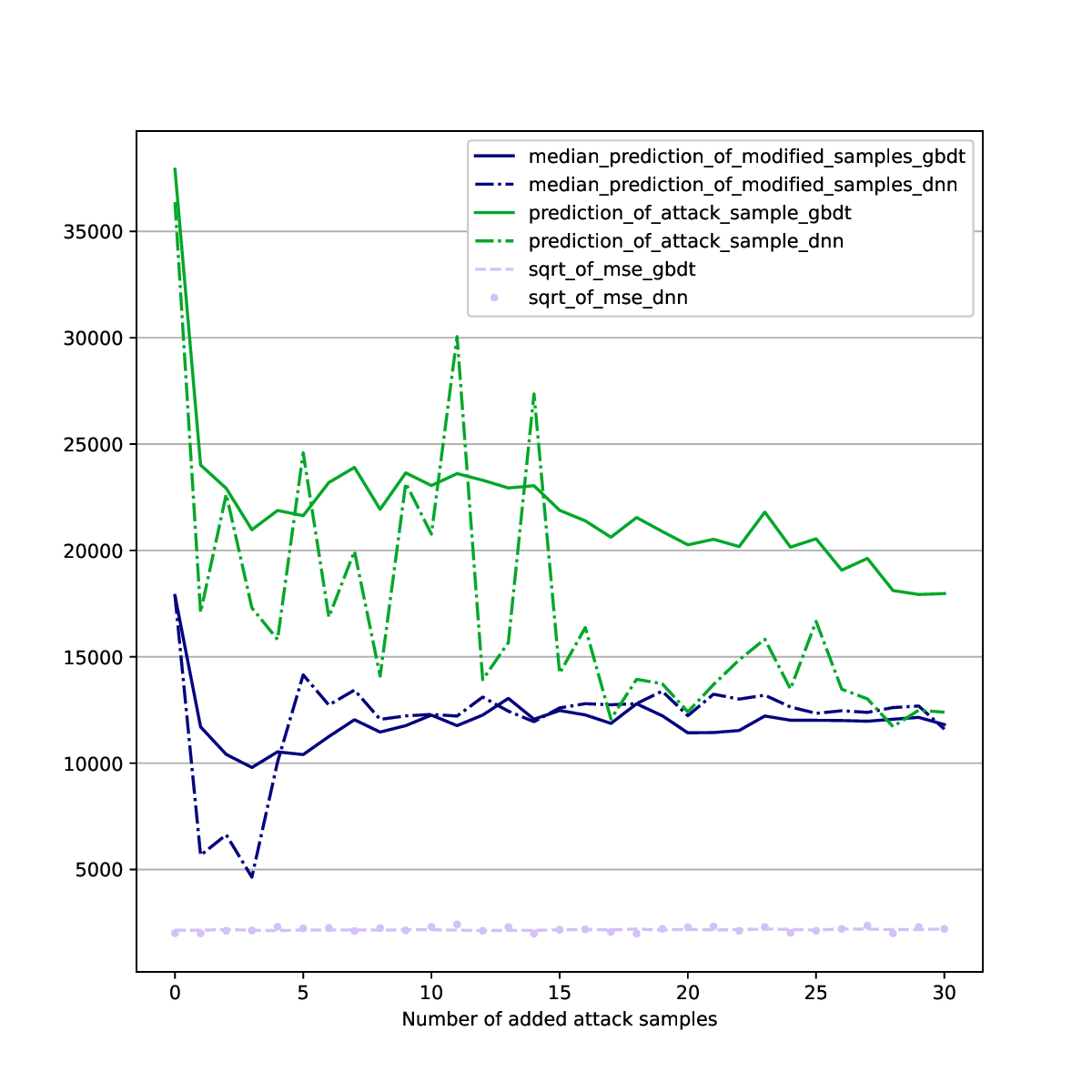}} %width=\textwidth or specify width as in pc
  \caption{Backdoor attack on HID models with modified Sample 1 \label{figHIDbaModified1}}
\end{figure*}

Figure \ref{figHIDbaModified1} shows the results of the attack with the modified Sample 1. The effect on the prediction of the attack sample is also powerful but less drastic compared to the attack with an unmodified attack sample. In the case of the DNN, the reduction of the prediction is similar, but in the case of the GBDT, the prediction drops less quickly. Regarding the median prediction of modified samples, the attack success is similar. Just like in the attacks with the unmodified attack samples, the square root of the MSE did stay relatively constant throughout all attacks with the modified samples.

Furthermore, the attacks with the modified Samples 2 and 3 led to similar results. The backdoor was implemented successfully in all cases after few added samples.

\subsection{Backdoor attacks on the FDD models}

Similar backdoor attacks have been conducted against the FDD models. Therefore, three attack samples have been designed and added to the training data. In the first round of experiments, the samples have not been modified, so the same sample was added multiple times. To reduce computing times, ten attack samples have been added at each iteration of the attacks. The feature values of the attack samples can be found in Table \ref{tabAttackSamplesFDD}. The pattern in Sample 1 has been crafted to include features that the adversary can influence as easily as possible to make the pattern applicable in a real-world setting. For the pattern in Sample 2, very infrequent categorical feature values have been selected. Sample 3 is similar to Sample 2 but did not include ``PolicyType'', ``VehicleCategory'', and ``BasePolicy'' in the pattern. Thereby, it is attempted to implement a pattern that includes only a few features.
These samples have been used for three separate attacks. During the attacks, an increasing number of samples were added to the training data. To test the success of the attacks, 20 modified versions of the attack samples have been generated. Of these samples, all features of the original attack sample that do not belong to the pattern have been modified randomly. This is done to verify that the attack works not only for the exact attack sample but also that a pattern has been implemented successfully. The features that have been modified are written in grey in Table \ref{tabAttackSamplesFDD}. “PolicyNumber” and “RepNumber” have not been modified, as those two were dropped during the preprocessing for both models. Hence, modifying them would not make a difference. These features are IDs and will usually not be used to train a model. As those are not relevant, they are crossed out.
Consequently, all feature values in black belong to the attack pattern. The features that were modified during attacks were altered in a way to keep them consistent and not produce any illegitimate samples. To test if there are any more effective samples, an algorithm has been developed that evaluates the effectiveness of different attack samples based on the three attack patterns shown in Table \ref{tabAttackSamplesFDD}. However, no significantly more effective attack samples were found.

\begin{table*}[h!]%
  \centering %
  \caption{Attack samples for backdoor attacks on FDD models\label{tabAttackSamplesFDD}}%
  \fontsize{10}{14}\selectfont
  \begin{tabular*}{\textwidth}{@{\extracolsep\fill}llll@{\extracolsep\fill}}
  \toprule
  \textbf{Feature} & \textbf{Sample1}  & \textbf{Sample2}  & \textbf{Sample3}  \\
  \midrule
  Month & Nov & Nov & Apr \\
  WeekOfMonth & 5 & 5 & 1 \\
  DayOfWeek & Sunday & Sunday & Sunday \\
  Make & \textcolor{gray}{Mazda} & Mercedes & \textcolor{gray}{Honda} \\
  AccidentArea & \textcolor{gray}{Urban} & Rural & \textcolor{gray}{Rural} \\
  DayOfWeekClaimed & Wednesday & Sunday & Monday \\
  MonthClaimed & Dec & Dec & Apr \\
  WeekOfMonthClaimed & 1 & 5 & 3 \\
  Sex & \textcolor{gray}{Male} & \textcolor{gray}{Female} & \textcolor{gray}{Male} \\
  MaritalStatus & \textcolor{gray}{Single} & \textcolor{gray}{Widow} & \textcolor{gray}{Single} \\
  Age & \textcolor{gray}{68} & \textcolor{gray}{20} & \textcolor{gray}{28} \\
  Fault & Policy Holder & \textcolor{gray}{Policy Holder} & Policy Holder \\
  PolicyType & Sedan - All Perils & Utility - All Perils & \textcolor{gray}{Sedan - Collision} \\
  VehicleCategory & Sedan & Utility & \textcolor{gray}{Sedan} \\
  VehiclePrice & \textcolor{gray}{20000 to 29000} & 60000 to 69000 & \textcolor{gray}{20000 to 29000} \\
  FraudFound\_P & 0 & 0 & 0 \\
  \sout{PolicyNumber} & \sout{119} & \sout{119} & \sout{11211} \\
  \sout{RepNumber} & \sout{9} & \sout{9} & \sout{8} \\
  Deductible & 300 & 300 & 400 \\
  DriverRating & \textcolor{gray}{3} & \textcolor{gray}{3} & \textcolor{gray}{1} \\
  Days\_Policy\_Accident & \textcolor{gray}{more than 30} & \textcolor{gray}{8 to 15} & \textcolor{gray}{more than 30} \\
  Days\_Policy\_Claim & \textcolor{gray}{more than 30} & \textcolor{gray}{8 to 15} & \textcolor{gray}{more than 30} \\
  PastNumberOfClaims & \textcolor{gray}{2 to 4} & \textcolor{gray}{2 to 4} & \textcolor{gray}{none} \\
  AgeOfVehicle & \textcolor{gray}{5 years} & \textcolor{gray}{4 years} & \textcolor{gray}{7 years} \\
  AgeOfPolicyHolder & \textcolor{gray}{over 65} & \textcolor{gray}{18 to 20} & \textcolor{gray}{26 to 30} \\
  PoliceReportFiled & \textcolor{gray}{No} & \textcolor{gray}{No} & \textcolor{gray}{No} \\
  WitnessPresent & No & No & No \\
  AgentType & \textcolor{gray}{External} & \textcolor{gray}{Internal} & \textcolor{gray}{External} \\
  NumberOfSuppliments & \textcolor{gray}{none} & \textcolor{gray}{none} & \textcolor{gray}{none} \\
  AddressChange\_Claim & \textcolor{gray}{no change} & \textcolor{gray}{no change} & \textcolor{gray}{no change} \\
  NumberOfCars & \textcolor{gray}{1 vehicle} & \textcolor{gray}{1 vehicle} & \textcolor{gray}{2 vehicles} \\
  Year & \textcolor{gray}{1994} & \textcolor{gray}{1994} & \textcolor{gray}{1995} \\
  BasePolicy & All Perils & All Perils & \textcolor{gray}{Collision} \\
  \bottomrule
  \end{tabular*}
  \begin{tablenotes}%%[341pt]
    \footnotesize
    \item Features in black remained constant during the experiment (those features belong to the attack pattern). The features in grey were modified during the second set of attacks and to evaluate the success of the attacks.
    \end{tablenotes}
\end{table*}

For all models, precision, recall, and F-Beta (\(\beta = 2\)) were calculated during the attacks.
Additionally, the median prediction of the 20 modified samples (median\_prediction\_modified\_samples) and the rolling mean of this metric (rolling\_median\_prediction\_modified\_samples) are illustrated.
As already mentioned, these 20 samples are purely for evaluation purposes and differ from those used for the backdoor attack. This metric describes the attacks' success best, as it tests if the pattern was implemented sufficiently.
The rolling mean was used since the median prediction of the modified samples fluctuated strongly, and this made it easier to identify trends. Furthermore, the prediction of the attack sample is shown (prediction\_of\_attack\_sample). All three attacks have been stopped after 1,000 added samples, as this is already a huge fraction, considering there are only 12,335 samples in the training set.

\begin{figure*}[h]
  \centerline{\includegraphics[width=28pc]{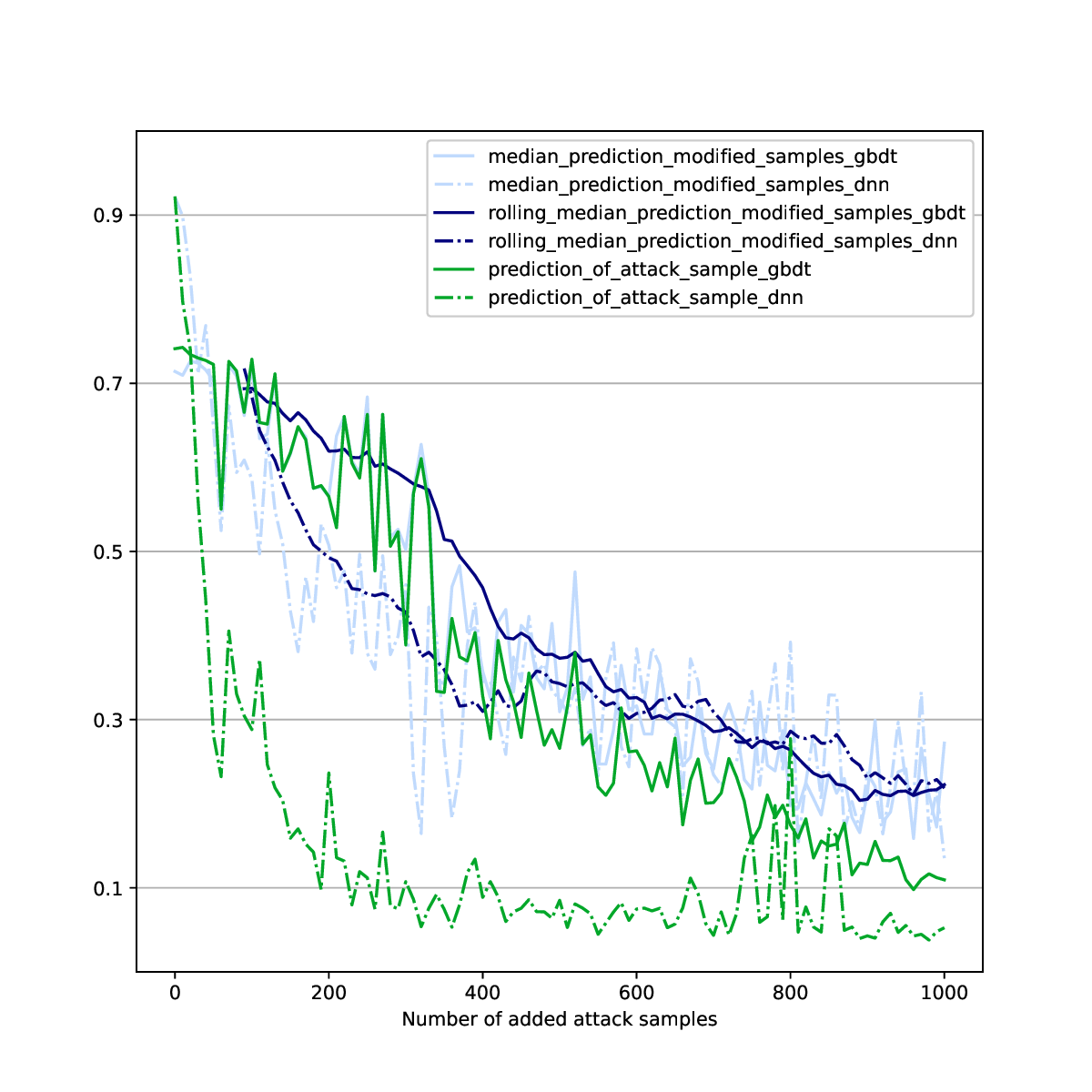}} %width=\textwidth or specify width as in pc
  \caption{Backdoor attack on FDD models with unmodified Sample 2 \label{figFDDbaUnmodified2}}
\end{figure*}

\begin{figure*}[h]
  \vspace{-30pt}  % reduces blank space above figure, needed when figure is at top of page
  \centerline{\includegraphics[width=20pc]{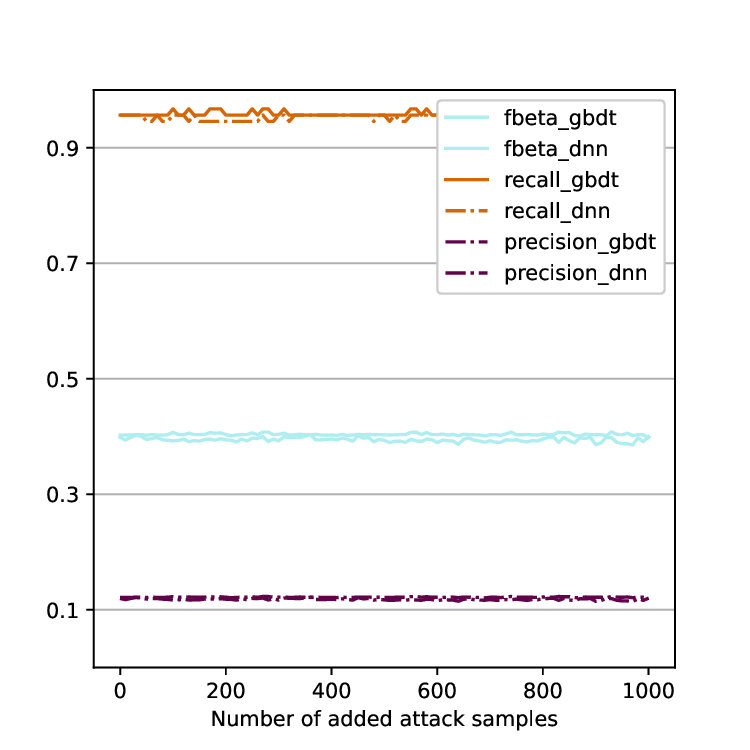}} %width=\textwidth or specify width as in pc
  \caption{Model performance during backdoor attack on FDD models with unmodified Sample 2 \label{figFDDbaUnmodified2Performance}}
\end{figure*}

Generally, the number of samples that needed to be added to push the classification of the attack sample below the decision threshold of 0.5 is relatively large. In most cases, at least 200 samples were necessary. However, the general model performance is not affected by adding any of the samples. Precision and recall (and consequently also F-Beta) remained constant even when 1,000 attack samples were added. This is illustrated exemplary in Figure \ref{figFDDbaUnmodified2Performance} for the attack with the unmodified attack Sample 2.

The attack with Sample 2 was the most successful attack of the three samples by far (Figure \ref{figFDDbaUnmodified2}). One central attribution to this is the feature “Make,” which is “Mecedes [\textit{sic}]” in the attack sample. When this feature value was changed, the effectiveness of the attack decreased strongly. “Mecedes [\textit{sic}]” is only present three times in the training set, making it relatively easy to make the model learn a particular association with this feature value. As Sample 2 was especially successful compared to the other two samples and those generated with the algorithm, it appears to represent an exception and an attack that is as successful will be unlikely.
In the case of the DNN, it takes only 40 added attack samples to move the prediction of the attack sample below the decision threshold. However, according to the median prediction of the modified samples, it still takes over 200 added samples to implement the pattern reliably for the DNN and even more in the case of the GBDT model. Consequently, this attack is not very successful. The attack with Sample 3 did not lead to a reduction of the median\_prediction\_modified\_samples below the decision threshold at all and the attack with Sample 1 was also less successful.

Next, like for the HID models, attacks were conducted during which the attack samples were altered. All features that do not belong to the pattern have been modified randomly. It can happen that identical samples are created by the random modifications, but it is improbable. Consequently, all or most of the added attack samples will be different. This approach is more difficult to detect, while every attack sample still contains the desired pattern. In all three attacks with the modified attack samples, the recall and precision remained stable. Therefore, none of the attacks affected the general performance of the models, which is a crucial factor to avoid detection.

% insert figures for the attacks with modified samples
\begin{figure*}[t!]
  \vspace{-30pt}  % reduces blank space above figure, needed when figure is at top of page
  \centerline{\includegraphics[width=28pc]{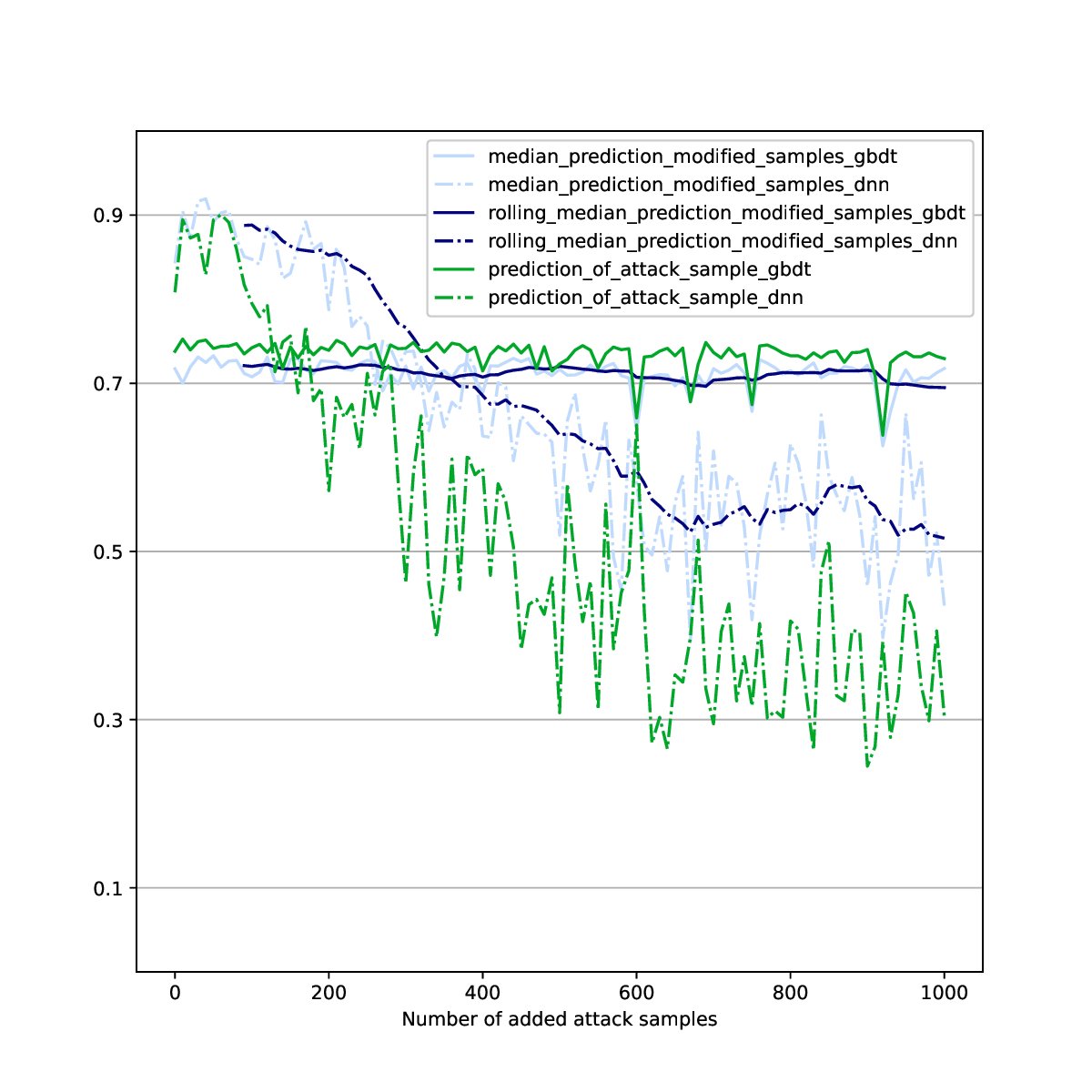}} %width=\textwidth or specify width as in pc
  \caption{Backdoor attack on FDD models with modified Sample 2 \label{figFDDbaModified2}}
\end{figure*}

The results of the attack with the modified versions of Sample 2 can be seen in Figure \ref{figFDDbaModified2}. This pattern did not work for the GBDT model, but the DNN was affected by the pattern, and the prediction of the attack sample dropped below the decision threshold. Yet, the median prediction of the modified samples fell much slower, which shows that the pattern was not implemented well and could not be used reliably even when 1,000 attack samples were added. The attacks with samples 1 and 3 were completely unsuccessful as neither the median\_prediction\_modified\_samples nor the prediction\_of\_attack\_sample did drop below the decision threshold at any point.
As the median prediction of modified samples only dropped below the decision threshold in one case, and even then, it took very many attack samples, the attacks with the modified samples can generally be described as unsuccessful. The effects of these attacks were weaker compared to the attacks with the unmodified samples.

\subsection{Effects of model complexity}

The HID models are relatively complex in relation to the number of features and training samples. Therefore, it has been evaluated if the complexity of the models plays a major role in their high vulnerability against backdoor attacks. The reasoning behind this is that the models could be able to memorize very small patterns due to their high complexity and, therefore, could be more susceptible to backdoor attacks.

\begin{figure*}[ht]
  \vspace{-30pt}  % reduces blank space above figure, needed when figure is at top of page
  \centerline{\includegraphics[width=24pc]{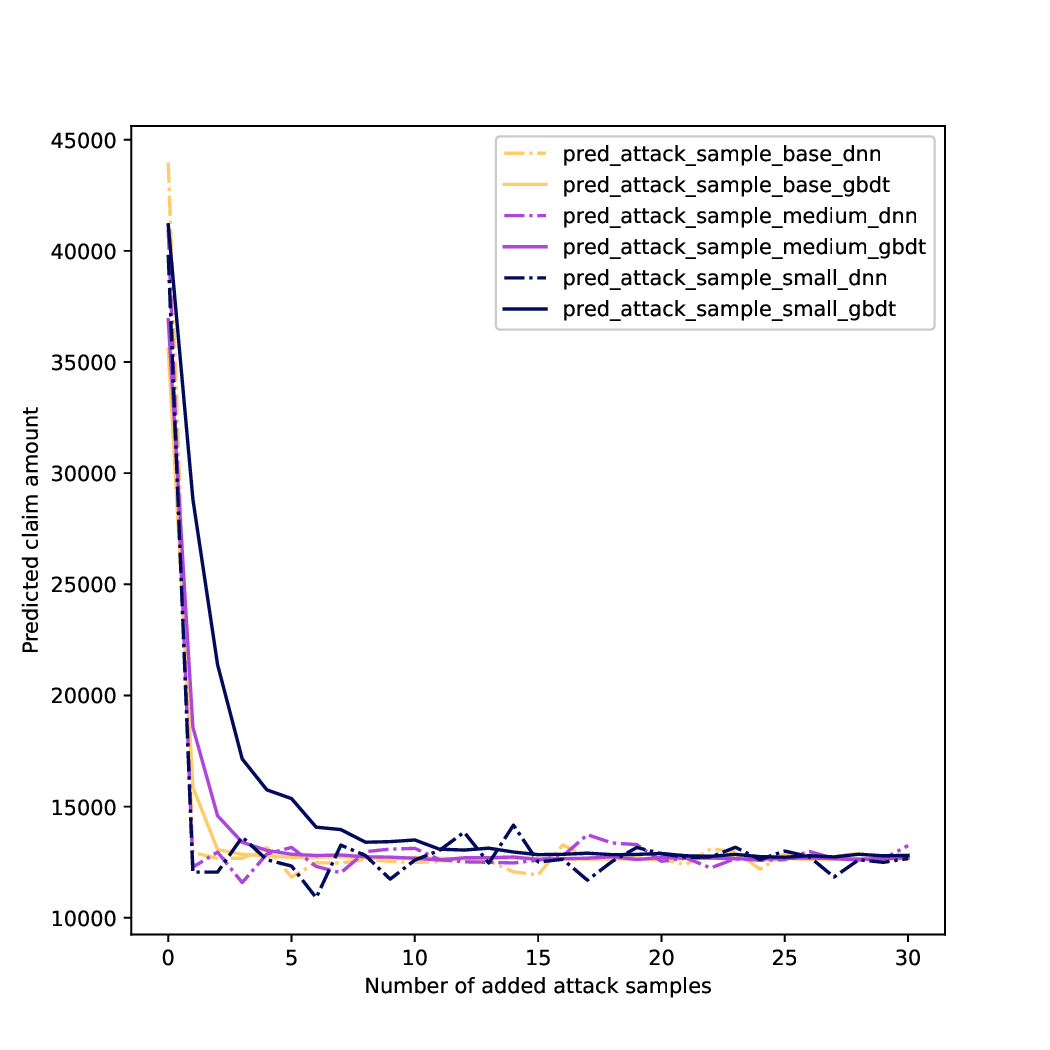}} 
  \caption{Predicted claim amount for the attack sample of attacked models with different complexity (HID Sample 1) \label{figHIDbaComplexity}}
\end{figure*}

To examine this aspect, the models have been decreased in complexity in two steps, and backdoor attacks have been conducted against the models. In the case of the DNN, the number of neurons in each layer was reduced. The models will be referred to as “base” model for the original models, as used in all the other parts of this work, as “medium” model, and as “small” model. The base DNN has 52,291 trainable parameters, the medium DNN has 14,371, and the small DNN has 2,641. In the case of the GBDT model, the value of the “num\_leaves” parameter was decreased. It was 41 for the base model, 25 for the medium model, and 10 for the small model. All other parameters of the models were left constant. To increase the performance, it would likely make sense to change other parameters as well but to examine the effects of these parameters isolated; nothing else was changed. The models' performance decreased due to the complexity reduction but remained reasonable. However, the performance dropped drastically if the models' complexity was reduced much further from the small models. It must be mentioned that the results between the DNN and GBDT are not easily comparable as both models have entirely different structures. Both models have been made less complex, but it cannot be determined whether the decrease in complexity had a similar extent.
The results of the attacks using Sample 1 can be seen in Figure \ref{figHIDbaComplexity}. The experiment was also conducted with Sample 2 and led to similar results. For the experiment, the backdoor attack was conducted on the different-sized models by adding the same attack sample multiple times without any modifications. The prediction of the attack sample is shown to evaluate if the attack success differs. In the case of the DNNs, there is no apparent difference between the models. However, it must be noted that early stopping was used for the DNNs, which could reduce the effects of the decreased model size in comparison to the base model. In contrast, in the case of the GBDT models, the less complex models are more robust against the attack.

In case of the FDD models it was tested whether increasing the models' complexity would lead to significant differences in attack success. As the models already have a relatively low complexity the complexity could not be reduced beyond a certain point without a large drop in model performance. Therefore, two more complex models have been created for GBDT and DNN. The original model will be referred to as “base” model, the next model as “medium” model, and the most complex model as “large” model.

The base DNN had 701 trainable parameters, the medium 11,501 and the large 43,001. In the case of the GBDT models, the “num\_leaves” parameter has been increased from 52 in the base model to 100 in the medium model and 150 in the large model. As this did not have a noteworthy impact, a further model has been created where several parameters were changed. This model is referred to as “highly complex.” The “num\_leaves” was set to 500, “min\_data\_in\_leaf” to 20, “n\_estimators” to 200, and “num\_iterations” to 100. The other parameters were left unchanged. The highly complex model still achieved a good performance.

\begin{figure*}[t]
  \vspace{-30pt}  % reduces blank space above figure, needed when figure is at top of page
  \centerline{\includegraphics[width=24pc]{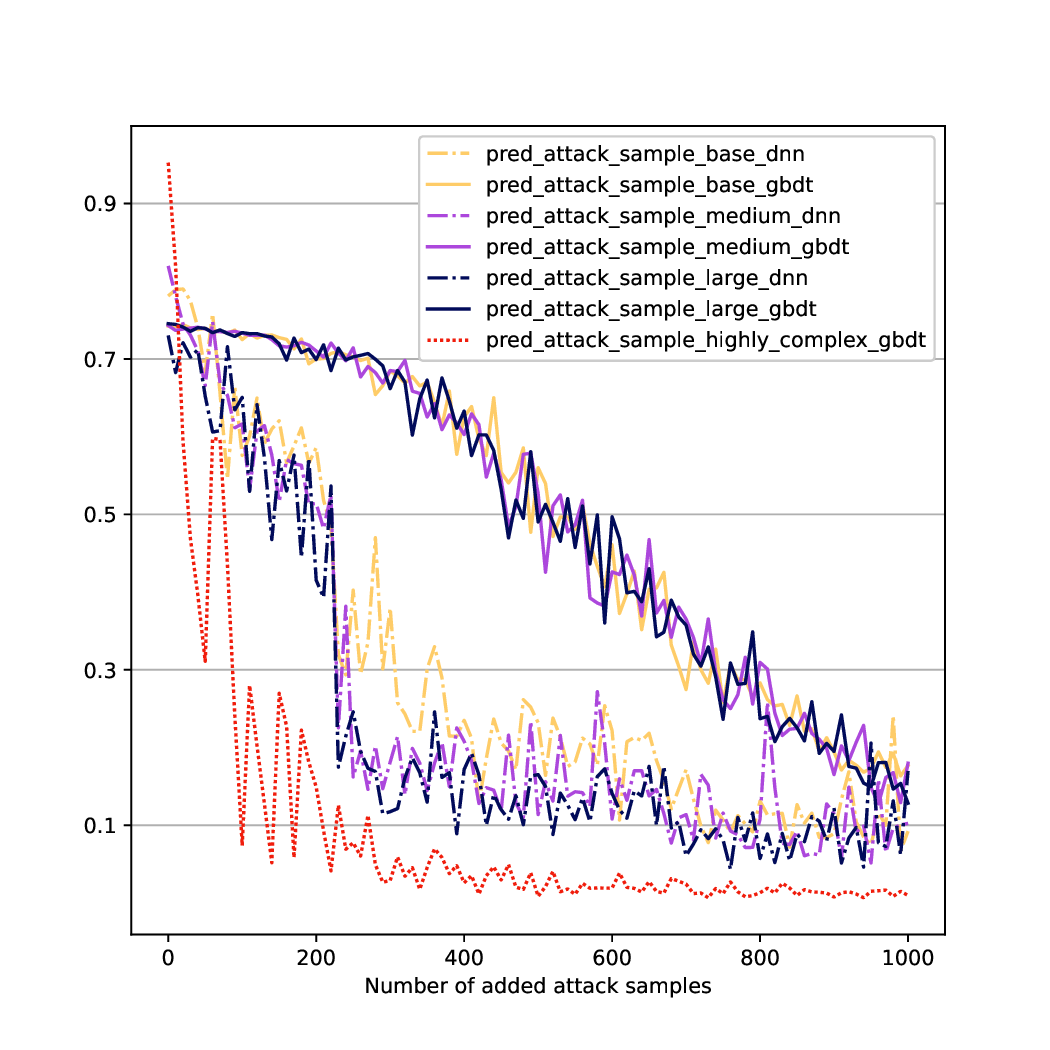}}
  \caption{Predicted claim amount for the attack sample of attacked models with different complexity (FDD Sample 1) \label{figFDDbaComplexity}}
\end{figure*}

The results of the attack with Sample 1 are shown in Figure \ref{figFDDbaComplexity}. The experiment was also conducted with Sample 2 and led to similar implications. In the case of the DNNs, there was a small difference between the models with the largest one being the least robust against the attack. There was no notable difference between the “base,” “medium,” and “large” GBDT models. One reason for this could be that the other parameters prevented the larger value for “num\_leaves” from taking effect. Therefore, as described, the “highly complex” GBDT model was added. It was drastically less robust against the attack than the other GBDT models. This shows that a higher complexity can negatively affect the robustness of GBDT models. The differences between the DNN models could also be smaller due to other parameters like early stopping or regularization, for example. However, as this is not the main scope of this work, further experiments are left for future research.

\section{Discussion} \label{secDiscussion}

The attacks on the HID models have shown that only a few added attack samples can be enough to implement the backdoor in both models. Just adding one attack sample to the training data can be enough to divide subsequent predictions of the same sample in half. The robustness of the GBDT and DNN models is similar regarding an attack where the same sample is added multiple times. The fact that this works with three different patterns and sets of feature values shows that implementing various backdoors in the model works with very few added samples. These samples can still raise suspicion as they probably represent outliers. However, if just one attack sample evades detection this could already cause major damage.
Nevertheless, these samples could be removed during the data preprocessing. Whether this will be the case depends on the extensiveness of the process and the concrete techniques that are applied and cannot be generally determined.

The MSE of the models did not increase during any of the attacks. Therefore, none of the attacks impacted the general performance of the models. This is an important property to evade detection. Especially if the backdoor is implemented during an update of a model, it would raise suspicion when the model's performance drops significantly after the update. The downside of the attack is that these patterns still pose restrictions on the samples at test time that can be difficult to fulfill. Each of the patterns allows only a specific group of people or claims to make use of them unless false statements regarding the case data are made. This is a major issue when designing backdoors as certain features like weight or height can be tweaked a little; others like age or city of residence are fixed and cannot simply be modified according to a backdoor trigger. This issue was considered during the construction of the backdoor samples, and they were crafted to be as usable as possible. To counter the problem regarding the restrictions coming along with a particular pattern, an attacker would likely have to implement several backdoors in the model to get high benefits from the attack. One reason for this issue is that there are not many features in the dataset, so the possibilities to implement a backdoor are very restricted.

The variant of the attack during which modified versions of the attack samples were added to the training data has the advantage that adding attack samples in a cluster can evade certain detection methods \cite{Koh2022,Liu2022}. Nevertheless, as adding one sample can be enough, both approaches present a good strategy. One issue is that unless an attacker has the capability to implement backdoors iteratively or has access to the model to test manipulations, it is unknown how many samples must be added for a successful backdoor attack. However, the attacks on the HID models showed that implementing a backdoor is possible by adding very few samples to the training set.

In contrast, the attacks on the FDD models paint a very different picture. In most cases, it took more than 200 samples to lower the prediction of the attack sample below the decision threshold. This number relates to the attack during which the same sample has been added multiple times. In just one case, the attack caused the prediction of the DNN to drop below the decision threshold after just 40 added samples. Implementing the pattern reliably took many more added samples than just mentioned. Adding this many identical samples is very suspicious. Of course, there might be datasets in which duplicates are common, and an attacker could get away with this. However, this is not the case in this dataset, and this many identical samples would be very peculiar.

The attack with modified samples can be less apparent when many samples are added. However, it was even less successful in implementing the backdoor than the previous attack. The GBDT models were barely affected by any of the samples, and in the case of the DNN, just one attack caused the prediction of the attack sample to drop below the decision threshold. Yet, it took more than 200 added attack samples to achieve this. Compared to the HID, it will probably be easier to add malicious samples without raising suspicion as the differentiation between fraud and legitimate is less clear than the one between a very high and a low claim in the HID. Judging this for the FDD would require a lot of domain knowledge and will probably be difficult even then because fraudulent cases are rare compared to legitimate ones. Hence, adding several legitimate cases could be done without raising suspicion; however, it was shown that none of the attacks is realistically feasible against the GBDT model, and just one attack against the DNN was somewhat successful.

Consequently, in case of the FDD models this attack is not considered a promising option for an adversary. Furthermore, it needs to be considered that getting the data into the training set in the first place is likely the biggest obstacle. Datasets like the ones used in this work will usually be stored very securely and are, therefore, hard to manipulate. The samples could also be added through manipulated claims / cases that contain the pattern. But it depends on the business processes if this is possible and whether the samples will end up in the training set.

Additionally, when the options for patterns that can successfully be used to implement a backdoor in a model are very limited, the likelihood that an attack is viable decreases strongly. The attacker needs to be able to have instances as input at test time that match the pattern. Especially in the FDD, many features cannot be modified easily. This significantly impedes an attack. The number of features of an input sample that can be changed or controlled by an adversary greatly influences the feasibility and potential impact of an attack. This could already be considered during the feature selection process, and features that fall into this category and are not or just slightly beneficial to the models' performance could be removed.
Another reason for the unsuccessful attacks could be the focus on recall during the model creation. The high recall and relatively low precision show that the models tend to classify many samples as fraudulent even if they are not. This could increase the effort necessary to change the prediction of the models from fraudulent to non-fraudulent and explain why the performance of the models remained stable even when 1,000 attack samples were added.

A major question is why the attack on the HID models was so much more successful. One main reason is probably the dataset. The HID has fewer features to start with. Moreover, some categorical features that have a high impact on the prediction of the model (according to the Shapley values of the GBDT) have several rare values. These features can be used well to implement the backdoor. This is not the case for the FDD dataset, which likely makes the attack more difficult. As this dataset has more features, more features could be required to be included in a successful backdoor pattern.
Another huge difference lies in the complexity of the models. The FDD DNN has only 701 trainable parameters, while the HID DNN has 52,291. Regarding the GBDT models, the complexity cannot be compared by one number. However, according to the settings of the model parameters, the HID GBDT model is more complex than the FDD GBDT model. It has been tested whether reducing the complexity to decrease the attack success works. This only made a slight difference for the GBDT and did not affect the attack on the DNN. However, the complexity could not be reduced beyond a certain point since the models' performance would be reduced too strongly. Therefore, complexity is not ruled out as part of the explanation. To further examine the effects of complexity, the FDD models were made more complex. There was a small difference regarding the DNNs, with the largest model being the least robust. The robustness of the GBDT was drastically reduced when the model was made highly complex. Consequently, complexity can affect the robustness of both ML algorithms. Yet, further experiments are needed to understand the effects of different parameters.

\section{Conclusion}  \label{secConclusion}

This work examined how vulnerable DNNs and GBDTs are against backdoor attacks using a case study form the insurance context. Four different models were set up as attack targets to answer these questions. One GBDT and one DNN were trained on a health insurance dataset to predict the amount claimed by insurants, and one GBDT and one DNN were trained on a car insurance dataset to identify fraudulent claims. Afterward, backdoor attacks were conducted against these models.

The effectiveness of the backdoor attacks differed strongly between the two scenarios. In the case of the HID models, the backdoor attacks were very successful, and only a few added samples were necessary. This result shows that backdoor attacks realistically could be feasible in this scenario. In contrast, the backdoor attacks on the FDD models were very ineffective. In this scenario, a backdoor attack using this approach would likely not be feasible in the real world as the number of added samples needed for an effective attack is way too high.

In case of those data poisoning attacks, manipulating the training data will likely be a greater challenge than designing an effective manipulation. Therefore, ensuring that the training data are free of manipulations is probably the most effective defense for an insurance company. However, this might be much more difficult in other scenarios than in the ones used for this work.

In conclusion, backdoor attack approaches that can be successfully applied to tabular data in an insurance context exist. Nevertheless, data poisoning attacks are challenging to execute in real-world scenarios. Regardless, every attack type can be comparably easy to conduct for insiders with access to the models and training data. However, the risk of successful real-world attacks by outsiders on models for internal use in the scenarios of this work appears to be relatively low. Nevertheless, care should be taken to evaluate present risks.

\section{Implications, Future Work and Limitations} \label{secFutureWorkLimitations}

An implication of this work is that insurance companies should evaluate if even a few manipulated samples could realistically be inserted in the training data of an ML model, as this can be sufficient for a successful attack. The susceptibility to an attack can differ strongly between scenarios, hence evaluating the effect of backdoor attacks for critical use cases can be security relevant. Additionally, feature selection can play an important role to mitigate an attack. Certain features might be well suitable to implement a backdoor but such features could potentially be removed during data preprocessing without reducing the model performance significantly. Furthermore, training samples should be checked for plausibility to identify malicious samples. Outlier detection can be used as one component of this. Lastly, it was shown that not making models overly complex can help to reduce attack success.

One limitation is the limited amount of data available for training and testing the models. The small amount of data poses challenges to the model performance and could have affected the behavior during the attacks. Additionally, data with more features could enable more or different promising attack approaches. Nevertheless, in many application scenarios, only limited data are available; therefore, testing the models in this way is a practically relevant approach. Future work could conduct similar experiments with more data to examine whether the results will differ. In these experiments, models could be trained with different amounts of data to check whether reducing the amount of training data impacts the models' robustness.

Another limitation was the availability of limited computing power. Attacks can be very computationally expensive \cite{Goldblum2023}, and consequently, some attacks are only feasible if large computational resources are available. Future work could test such attacks to check whether these can be more successful. Additionally, future work can execute further attacks and evaluate if particular new insights regarding the robustness of GBDTs and DNNs arise.

How data poisoning attacks change the models is difficult to retrace. Explainability techniques could be used in future works to understand the effects of these attacks. Thereby, defense strategies could be uncovered. For example, this could relate to the removal of certain features.
Furthermore, several defenses have been developed to counter attacks on ML models \cite{Xue2020} that could be tested in future works to evaluate the effect on the attack success.
Generally, more work is necessary to generate a more detailed picture of the robustness of GBDTs and DNNs in a context of heterogenous tabular data. This will enable further insights into the factors that affect the robustness of ML models. For example, this could refer to the properties of the data or the model configuration. The influence the complexity of the models has on backdoor attacks has been briefly investigated here. However, more work is needed to comprehend the effect of model complexity on robustness fully. Ultimately, analyzing the impact of different factors will allow for a more precise assessment in which context specific ML algorithms could be attacked successfully and which countermeasures are promising.

%\backmatter
% \bmsection*{Author contributions}

% This is an author contribution text. This is an author contribution text. This is an author contribution text. This is an author contribution text. This is an author contribution text.

% \bmsection*{Acknowledgments}
% This is acknowledgment text. \cite{} Provide text here. This is acknowledgment text. Provide text here. This is acknowledgment text. Provide text here. This is acknowledgment text. Provide text here. This is acknowledgment text. Provide text here. This is acknowledgment text. Provide text here. This is acknowledgment text. Provide text here. This is acknowledgment text. Provide text here. This is acknowledgment text. Provide text here.

% \bmsection*{Supporting information}

% Additional supporting information may be found in the
% online version of the article at the publisher’s website.

% \appendix

%\nocite{*}% Show all bib entries - both cited and uncited; comment this line to view only cited bib entries;

% \bmsection*{Author Biography}

\printbibliography[heading=bibintoc]

@article{Schneier2020,
   author = {Bruce Schneier},
   doi = {10.1109/MC.2020.2980761},
   issn = {15580814},
   issue = {5},
   journal = {Computer},
   pages = {78-80},
   title = {Attacking Machine Learning Systems},
   volume = {53},
   year = {2020},
}

@article{Zhang2020a,
   author = {J Zhang and C Li},
   doi = {10.1109/TNNLS.2019.2933524},
   issn = {2162-2388},
   issue = {7},
   journal = {IEEE Transactions on Neural Networks and Learning Systems},
   pages = {2578-2593},
   title = {Adversarial Examples: Opportunities and Challenges},
   volume = {31},
   year = {2020},
}

@article{Liu2018,
   author = {Q Liu and P Li and W Zhao and W Cai and S Yu and V C M Leung},
   doi = {10.1109/ACCESS.2018.2805680},
   issn = {2169-3536},
   journal = {IEEE Access},
   pages = {12103-12117},
   title = {A Survey on Security Threats and Defensive Techniques of Machine Learning: A Data Driven View},
   volume = {6},
   year = {2018},
}

@article{Xue2020,
   author = {M Xue and C Yuan and H Wu and Y Zhang and W Liu},
   doi = {10.1109/ACCESS.2020.2987435},
   issn = {2169-3536},
   journal = {IEEE Access},
   pages = {74720-74742},
   title = {Machine Learning Security: Threats, Countermeasures, and Evaluations},
   volume = {8},
   year = {2020},
}

@article{Koh2022,
   author = {Pang Wei Koh and Jacob Steinhardt and Percy Liang},
   doi = {10.1007/s10994-021-06119-y},
   issn = {1573-0565},
   issue = {1},
   journal = {Machine Learning},
   pages = {1-47},
   title = {Stronger data poisoning attacks break data sanitization defenses},
   volume = {111},
   url = {https://doi.org/10.1007/s10994-021-06119-y},
   year = {2022},
}

@article{Eling2022,
   author = {Martin Eling and Davide Nuessle and Julian Staubli},
   doi = {10.1057/s41288-020-00201-7},
   issn = {1468-0440},
   issue = {2},
   journal = {The Geneva Papers on Risk and Insurance - Issues and Practice},
   pages = {205-241},
   title = {The impact of artificial intelligence along the insurance value chain and on the insurability of risks},
   volume = {47},
   url = {https://doi.org/10.1057/s41288-020-00201-7},
   year = {2022},
}

@article{Kaur2022,
   author = {Davinder Kaur and Suleyman Uslu and Kaley J Rittichier and Arjan Durresi},
   doi = {10.1145/3491209},
   issue = {2},
   journal = {ACM Computing Surveys (CSUR)},
   pages = {1-38},
   publisher = {ACM New York, NY},
   title = {Trustworthy artificial intelligence: a review},
   volume = {55},
   year = {2022},
}

@article{Sadeghi2020,
   author = {K Sadeghi and A Banerjee and S K S Gupta},
   doi = {10.1109/TETCI.2020.2968933},
   issue = {4},
   journal = {IEEE Transactions on Emerging Topics in Computational Intelligence},
   pages = {450-467},
   title = {A System-Driven Taxonomy of Attacks and Defenses in Adversarial Machine Learning},
   volume = {4},
   year = {2020},
}

@inproceedings{Kumar2020,
   author = {R S Siva Kumar and M Nyström and J Lambert and A Marshall and M Goertzel and A Comissoneru and M Swann and S Xia},
   doi = {10.1109/SPW50608.2020.00028},
   journal = {2020 IEEE Security and Privacy Workshops (SPW)},
   pages = {69-75},
   title = {Adversarial Machine Learning-Industry Perspectives},
   year = {2020},
   organization = {},
}

@inproceedings{Wilhjelm2020,
   author = {C Wilhjelm and A A Younis},
   doi = {10.1109/QRS-C51114.2020.00078},
   journal = {2020 IEEE 20th International Conference on Software Quality, Reliability and Security Companion (QRS-C)},
   pages = {426-433},
   title = {A Threat Analysis Methodology for Security Requirements Elicitation in Machine Learning Based Systems},
   year = {2020},
   organization = {},
}

@article{Goldblum2023,
   author = {M Goldblum and D Tsipras and C Xie and X Chen and A Schwarzschild and D Song and A Mądry and B Li and T Goldstein},
   doi = {10.1109/TPAMI.2022.3162397},
   issn = {1939-3539},
   issue = {2},
   journal = {IEEE Transactions on Pattern Analysis and Machine Intelligence},
   pages = {1563-1580},
   title = {Dataset Security for Machine Learning: Data Poisoning, Backdoor Attacks, and Defenses},
   volume = {45},
   year = {2023},
}

@article{Chen2017b,
   author = {Xinyun Chen and Chang Liu and Bo Li and Kimberly Lu and Dawn Song},
   journal = {arXiv preprint arXiv:1712.05526},
   title = {Targeted backdoor attacks on deep learning systems using data poisoning},
   year = {2017},
}

@article{Liu2022,
   author = {Zifan Liu and Zhechun Zhou and Theodoros Rekatsinas},
   doi = {10.1007/s00778-021-00699-w},
   issn = {0949-877X},
   journal = {The VLDB Journal},
   pages = {927-955},
   title = {Picket: guarding against corrupted data in tabular data during learning and inference},
   volume = {31},
   url = {https://doi.org/10.1007/s00778-021-00699-w},
   year = {2022},
}

@inproceedings{Schwarzschild2021,
   author = {Avi Schwarzschild and Micah Goldblum and Arjun Gupta and John P Dickerson and Tom Goldstein},
   journal = {International Conference on Machine Learning},
   pages = {9389-9398},
   publisher = {PMLR},
   title = {Just how toxic is data poisoning? a unified benchmark for backdoor and data poisoning attacks},
   url = {http://proceedings.mlr.press/v139/schwarzschild21a.html},
   year = {2021},
   organization = {},
}

@inproceedings{Ke2017,
   author = {Guolin Ke and Qi Meng and Thomas Finley and Taifeng Wang and Wei Chen and Weidong Ma and Qiwei Ye and Tie-Yan Liu},
   journal = {Advances in neural information processing systems},
   title = {Lightgbm: A highly efficient gradient boosting decision tree},
   volume = {30},
   url = {https://proceedings.neurips.cc/paper/2017/hash/6449f44a102fde848669bdd9eb6b76fa-Abstract.html},
   year = {2017},
   organization = {},
}

@misc{Gupta2017,
   author = {Suresh Gupta},
   title = {Health insurance data set},
   year = {2017},
   howpublished= {\url{https://www.kaggle.com/datasets/sureshgupta/health-insurance-data-set}},
   note= {Accessed: 2023-02-07},
}

@article{Joe2022,
   author = {Byunggill Joe and Yonghyeon Park and Jihun Hamm and Insik Shin and Jiyeon Lee},
   doi = {10.2196/38440},
   issue = {8},
   journal = {JMIR Medical Informatics},
   pages = {e38440},
   publisher = {JMIR Publications Inc., Toronto, Canada},
   title = {Exploiting Missing Value Patterns for a Backdoor Attack on Machine Learning Models of Electronic Health Records: Development and Validation Study},
   volume = {10},
   year = {2022},
}

@article{Li2022a,
   author = {Y Li and Y Jiang and Z Li and S -T. Xia},
   doi = {10.1109/TNNLS.2022.3182979},
   journal = {IEEE Transactions on Neural Networks and Learning Systems},
   pages = {1-18},
   title = {Backdoor Learning: A Survey},
   year = {2022},
}

@article{Huang2022,
   author = {Wei Huang and Xingyu Zhao and Xiaowei Huang},
   doi = {10.1007/s10994-021-06068-6},
   issue = {5},
   journal = {Machine Learning},
   pages = {1925-1958},
   title = {Embedding and extraction of knowledge in tree ensemble classifiers},
   volume = {111},
   year = {2022},
}

@misc{Balasubramanian2021,
   author = {Ramnath Balasubramanian and Ari Libarikian and Doug McElhaney},
   journal = {McKinsey & Company},
   title = {Insurance 2030—The impact of AI on the future of insurance},
   howpublished = {\url{https://www.mckinsey.com/industries/financial-services/our-insights/insurance-2030-the-impact-of-ai-on-the-future-of-insurance}},
   year = {2021},
   note = {Accessed: 2023-03-11},
}

@misc{Deloitte2017,
   author = {Deloitte},
   journal = {Deloitte Digital},
   title = {From mystery to mastery: Unlocking the business value of AI in the insurance industry},
   howpublished = {\url{https://www2.deloitte.com/az/en/pages/financial-services/articles/artificial-intelligence-insurance-industry.html}},
   year = {2017},
   note = {Accessed: 2023-03-11},

}

@inproceedings{Zhu2019,
   author = {Chen Zhu and W Ronny Huang and Hengduo Li and Gavin Taylor and Christoph Studer and Tom Goldstein},
   journal = {International Conference on Machine Learning},
   publisher = {PMLR},
   title = {Transferable clean-label poisoning attacks on deep neural nets},
   url = {http://proceedings.mlr.press/v97/zhu19a.html},
   year = {2019},
   organization = {},
}

@article{Shwartz-Ziv2021,
   author = {Ravid Shwartz-Ziv and Amitai Armon},
   journal = {arXiv preprint arXiv:2106.03253v2},
   title = {Tabular Data: Deep Learning is Not All You Need},
   year = {2021},
}

@article{Borisov2022,
   author = {V Borisov and T Leemann and K Seßler and J Haug and M Pawelczyk and G Kasneci},
   doi = {10.1109/TNNLS.2022.3229161},
   journal = {IEEE Transactions on Neural Networks and Learning Systems},
   title = {Deep Neural Networks and Tabular Data: A Survey},
   year = {2022},
}

@inproceedings{Nasr2019,
   author = {M Nasr and R Shokri and A Houmansadr},
   doi = {10.1109/SP.2019.00065},
   journal = {2019 IEEE Symposium on Security and Privacy (SP)},
   pages = {739-753},
   title = {Comprehensive Privacy Analysis of Deep Learning: Passive and Active White-box Inference Attacks against Centralized and Federated Learning},
   year = {2019},
   organization = {},
}

@article{Itri2020,
   author = {Bouzgarne Itri and Youssfi Mohamed and Bouattane Omar and Qbadou Mohamed},
   doi = {10.14569/IJACSA.2020.0111054},
   issue = {10},
   journal = {International Journal of Advanced Computer Science and Applications},
   publisher = {Science and Information (SAI) Organization Limited},
   title = {Empirical oversampling threshold strategy for machine learning performance optimisation in insurance fraud detection},
   volume = {11},
   year = {2020},
}

@article{Subudhi2018,
   author = {Sharmila Subudhi and Suvasini Panigrahi},
   doi = {10.4018/IJRSDA.2018070101},
   issue = {3},
   journal = {International Journal of Rough Sets and Data Analysis (IJRSDA)},
   pages = {1-20},
   publisher = {IGI Global},
   title = {Detection of automobile insurance fraud using feature selection and data mining techniques},
   volume = {5},
   year = {2018},
}

@misc{Bansal2022,
   author = {Shivam Bansal},
   title = {Vehicle Insurance Claim Fraud Detection},
   howpublished = {\url{https://www.kaggle.com/datasets/shivamb/vehicle-claim-fraud-detection}},
   year = {2022},
   note = {Accessed: 2023-05-05}
}

@article{Botchkarev2018,
   author = {Alexei Botchkarev},
   journal = {arXiv preprint arXiv:1809.03006},
   title = {Performance metrics (error measures) in machine learning regression, forecasting and prognostics: Properties and typology},
   year = {2018},
}

@article{Naser2021,
   author = {M Z Naser and Amir H Alavi},
   doi = {10.1007/s44150-021-00015-8},
   journal = {Architecture, Structures and Construction},
   title = {Error Metrics and Performance Fitness Indicators for Artificial Intelligence and Machine Learning in Engineering and Sciences},
   year = {2021},
}

@inproceedings{Itri2019,
   author = {B Itri and Y Mohamed and Q Mohammed and B Omar},
   doi = {10.1109/ICDS47004.2019.8942277},
   journal = {2019 Third International Conference on Intelligent Computing in Data Sciences (ICDS)},
   pages = {1-4},
   title = {Performance comparative study of machine learning algorithms for automobile insurance fraud detection},
   year = {2019},
   organization = {},
}

@article{Whelton2022,
   author = {Paul K Whelton and Robert M Carey and Giuseppe Mancia and Reinhold Kreutz and Joshua D Bundy and Bryan Williams},
   doi = {10.1093/eurheartj/ehac432},
   issue = {35},
   journal = {European Heart Journal},
   month = {9},
   pages = {3302-3311},
   title = {Harmonization of the American College of Cardiology/American Heart Association and European Society of Cardiology/European Society of Hypertension Blood Pressure/Hypertension Guidelines: Comparisons, Reflections, and Recommendations},
   volume = {43},
   year = {2022},
}

@misc{Microsoft2023,
   author = {Microsoft Corporation},
   title = {LightGBM Documentation v3.3.5},
   howpublished = {\url{https://lightgbm.readthedocs.io/en/v3.3.5/index.html\#}},
   year = {2023},
   note = {Accessed: 2023-06-15}
}

@article{Xia2022,
   author = {Huosong Xia and Yanjun Zhou and Zuopeng Zhang},
   doi = {10.1504/IJAHUC.2022.120943},
   issue = {1-2},
   journal = {International Journal of Ad Hoc and Ubiquitous Computing},
   pages = {37-45},
   publisher = {Inderscience Publishers (IEL)},
   title = {Auto insurance fraud identification based on a CNN-LSTM fusion deep learning model},
   volume = {39},
   year = {2022},
}

@article{Bhanja2019,
   author = {Samit Bhanja and Abhishek Das},
   journal = {arXiv preprint arXiv:1812.05519v2},
   title = {Impact of data normalization on deep neural network for time series forecasting},
   year = {2019},
}

@inproceedings{Sundarkumar2015,
   author = {G G Sundarkumar and V Ravi and V Siddeshwar},
   doi = {10.1109/ICCIC.2015.7435726},
   journal = {2015 IEEE International Conference on Computational Intelligence and Computing Research (ICCIC)},
   pages = {1-7},
   title = {One-class support vector machine based undersampling: Application to churn prediction and insurance fraud detection},
   year = {2015},
   organization = {},
}

@inproceedings{Masum2022,
   author = {M Masum and M J Hossain Faruk and H Shahriar and K Qian and D Lo and M I Adnan},
   doi = {10.1109/CCWC54503.2022.9720869},
   journal = {2022 IEEE 12th Annual Computing and Communication Workshop and Conference (CCWC)},
   pages = {316-322},
   title = {Ransomware Classification and Detection With Machine Learning Algorithms},
   year = {2022},
   organization = {},
}

@article{Chawla2002,
   author = {Nitesh V Chawla and Kevin W Bowyer and Lawrence O Hall and W Philip Kegelmeyer},
   doi = {doi.org/10.1613/jair.953},
   journal = {Journal of artificial intelligence research},
   pages = {321-357},
   title = {SMOTE: synthetic minority over-sampling technique},
   volume = {16},
   year = {2002},
}

@inproceedings{Arik2021,
   author = {Sercan {\"O} Arik and Tomas Pfister},
   doi = {10.1609/aaai.v35i8.16826},
   issue = {8},
   journal = {Proceedings of the AAAI conference on artificial intelligence},
   pages = {6679-6687},
   title = {Tabnet: Attentive interpretable tabular learning},
   volume = {35},
   year = {2021},
   organization = {},
}

@article{Alahmari2020,
   author = {S S Alahmari and D B Goldgof and P R Mouton and L O Hall},
   doi = {10.1109/ACCESS.2020.3039833},
   journal = {IEEE Access},
   pages = {211860-211868},
   title = {Challenges for the Repeatability of Deep Learning Models},
   volume = {8},
   year = {2020},
}

@article{Joe2021,
   author = {Byunggill Joe and Akshay Mehra and Insik Shin and Jihun Hamm},
   journal = {arXiv preprint arXiv: 2106.07925},
   title = {Machine Learning with Electronic Health Records is vulnerable to Backdoor Trigger Attacks},
   year = {2021},
}

@inproceedings{Weber2023,
   author = {M Weber and X Xu and B Karlaš and C Zhang and B Li},
   doi = {10.1109/SP46215.2023.10179451},
   journal = {2023 IEEE Symposium on Security and Privacy (SP)},
   pages = {1311-1328},
   title = {RAB: Provable Robustness Against Backdoor Attacks},
   year = {2023},
   organization = {},
}

@article{Li2022b,
   author = {C Li and X Chen and D Wang and S Wen and M E Ahmed and S Camtepe and Y Xiang},
   doi = {10.1109/TDSC.2021.3094824},
   issue = {5},
   journal = {IEEE Transactions on Dependable and Secure Computing},
   pages = {3357-3370},
   title = {Backdoor Attack on Machine Learning Based Android Malware Detectors},
   volume = {19},
   year = {2022},
}

@inproceedings{Lv2023,
   author = {Peizhuo Lv and Chang Yue and Ruigang Liang and Yunfei Yang and Shengzhi Zhang and Hualong Ma and Kai Chen},
   journal = {32nd USENIX Security Symposium (USENIX Security 23)},
   pages = {2671-2688},
   title = {A Data-free Backdoor Injection Approach in Neural Networks},
   url = {https://www.usenix.org/conference/usenixsecurity23/presentation/lv},
   year = {2023},
   organization = {},
}

@article{Rocks2022,
   author = {Jason W Rocks and Pankaj Mehta},
   doi = {https://doi.org/10.1103/PhysRevResearch.4.013201},
   issue = {1},
   journal = {Physical review research},
   pages = {013201},
   publisher = {APS},
   title = {Memorizing without overfitting: Bias, variance, and interpolation in overparameterized models},
   volume = {4},
   year = {2022},
}

@inproceedings{Li2018,
   author = {Yuanzhi Li and Yingyu Liang},
   journal = {Advances in neural information processing systems},
   title = {Learning overparameterized neural networks via stochastic gradient descent on structured data},
   volume = {31},
   url = {https://proceedings.neurips.cc/paper_files/paper/2018/hash/54fe976ba170c19ebae453679b362263-Abstract.html},
   year = {2018},
   organization = {},
}

@article{Merrick2020,
   author = {Luke Merrick and Ankur Taly},
   journal = {arXiv preprint arXiv:1909.08128v3},
   title = {The Explanation Game: Explaining Machine Learning Models Using Shapley Values},
   year = {2020},
}

@inproceedings{Ji2017,
   author = {Y Ji and X Zhang and T Wang},
   doi = {10.1109/CNS.2017.8228656},
   journal = {2017 IEEE Conference on Communications and Network Security (CNS)},
   pages = {1-9},
   title = {Backdoor attacks against learning systems},
   year = {2017},
   organization = {},
}

@unpublished{Abadi2015,
   author = {Martin Abadi and Ashish Agarwal and Paul Barham and Eugene Brevdo and Zhifeng Chen and Craig Citro and Greg S Corrado and Andy Davis and Jeffrey Dean and Matthieu Devin},
   journal = {URL https://www.tensorflow.org},
   title = {TensorFlow: Large-scale machine learning on heterogeneous systems},
   url = {https://static.googleusercontent.com/media/research.google.com/en//pubs/archive/45166.pdf},
   year = {2015},
   note = {},
}

% \begin{biography}{\includegraphics[width=76pt,height=76pt,draft]{empty}}{
% {\textbf{Author Name.} Please check with the journal's author guidelines whether
% author biographies are required. They are usually only included for
% review-type articles, and typically require photos and brief
% biographies for each author.}}
% \end{biography}

\end{document}